\pdfoutput=1
\documentclass{article}

\usepackage{microtype}
\usepackage{graphicx}
\usepackage{subcaption}
\usepackage{booktabs}
\usepackage{multirow}
\usepackage{hyperref}
\usepackage{algorithm}
\usepackage{algorithmic}

\usepackage[preprint]{icml2026}

\usepackage{amsmath,amssymb,mathtools,amsthm}
\usepackage{xcolor}
\usepackage[capitalize,noabbrev]{cleveref}

\theoremstyle{plain}

\theoremstyle{definition}

\icmltitlerunning{HexIt: Expert Iteration for Hexahedral Block Decomposition}

\begin{document}

\twocolumn[
  \icmltitle{HexIt: Learning Hexahedral Block Decomposition \\
             with Expert Iteration on Combinatorial Maps}

  \begin{icmlauthorlist}
    \icmlauthor{Arjun Narayanan}{ucb}
    \icmlauthor{Per-Olof Persson}{ucbmath}
  \end{icmlauthorlist}
  \icmlaffiliation{ucb}{University of California, Berkeley}
  \icmlaffiliation{ucbmath}{Department of Mathematics, University of California, Berkeley}
  \icmlcorrespondingauthor{Arjun Narayanan}{arjun.narayanan@berkeley.edu}
  \icmlkeywords{reinforcement learning, expert iteration, combinatorial maps, hexahedral meshing}
  \vskip 0.3in
]

\printAffiliationsAndNotice{}

\begin{abstract}
Decomposing a polyhedral solid into a few well-shaped hexahedral blocks is a
long-standing open problem in mesh generation. We cast it as a sequential
decision problem on a 3-dimensional generalised map: an agent splits edges,
faces and volumes with three insert operations until every cell is a
hexahedron and the mesh passes a quality threshold, and a policy that convolves
over the map's involutions chooses each insert. A successful decomposition
takes over a hundred decisions, and a uniformly random policy never finds one.
We clone a policy from constructed decompositions and improve it by expert
iteration. On held-out
chamfered solids that share no boundary topology with the cloning data, the
final policy solves an expected $94.4\%$ in $20$ samples, and expert iteration
raises the best-of-20 solve rate on large chamfered solids from $0.203$ to $0.355$. At
matched compute, PPO, GRPO and VinePPO updates all make the cloned policy
worse.

\end{abstract}

\section{Introduction}\label{sec:intro}

\begin{figure*}[t]
\centering
\includegraphics[width=\textwidth]{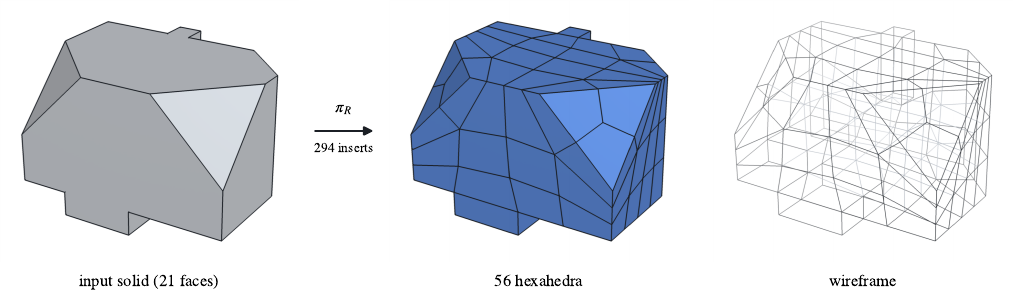}
\caption{A held-out large chamfered solid (large chamfered test set, $21$ faces; left) and the
all-hexahedral decomposition the final policy $\pi_R$ (seed 1) finds for it in
$294$ inserts: its $56$ hexahedra (middle) and the same mesh as a wireframe
(right). Its minimum scaled Jacobian is $0.50$, the solid's quality ceiling
$c(\Omega)$; $15$ of $60$ samples solve it.}
\label{fig:headline}
\end{figure*}

We study hexahedral block decomposition: partitioning a polyhedral solid into
a few well-shaped hexahedral blocks that meet face to face. Finite element
codes often
prefer hexahedral meshes to tetrahedral ones: tensor-product bases make element
computations cheaper, an advantage that grows with dimension; low-order
hexahedra are less prone to locking in structural mechanics; and boundary-layer
meshes are easier to build from layers of hexahedra
\citep{armstrong2015common,pietroni2022hex}. Producing a block decomposition
automatically is a long-standing open problem; even whether a quadrilateral boundary admits a
compatible hexahedral mesh is a global question
\citep{mitchell1996characterization,eppstein1999linear}.

We pose block decomposition as a sequential decision problem. The solid starts
as a single volume, and an agent edits it with three topological inserts
(split an edge at a new vertex, a face by a new edge, a volume by a new face on
a closed loop of existing edges) until every cell is a hexahedron and the
untangled mesh passes a quality threshold. The state is a 3-dimensional
generalised map (3-G-map; \citealp{damiand2014combinatorial}), which represents
every incidence as an involution on \emph{darts}, and the policy is a
convolution over these involutions. It sees no coordinates and its masks test
no geometry, so it is invariant to rigid motion and uniform scaling, and the
same action set applies to every polyhedron. This extends to three dimensions
work on planar quadrilateral block decompositions \citep{narayanan2026playing}.

Three features of this decision problem make it hard to learn. Episodes are
long: the median test solid that our policy solves takes $153$ decisions, and
one in ten takes more than $296$. Success is sparse and terminal, and whether a
partial decomposition can still be completed depends on global constraints
that show only late, so a policy that picks uniformly among legal moves never
succeeds on any of our test sets. And demonstrations exist only where a
decomposition can be written down, as for rectilinear solids and swept prisms,
not for solids whose edges and corners are chamfered by oblique planes.

We address these with expert iteration \citep{anthony2017thinking}. A policy
$\pi_0$ is cloned from \emph{certificates}, decompositions that are correct by
construction, turned into plans in the agent's action space by a backward
walk. Each round then samples the
current policy on a fixed set of mostly chamfered training solids that it does
not yet solve, keeps the best accepted decompositions of each solid in an
archive, and fine-tunes on the archive and the certificates.

\begin{figure*}[t]
\centering
\includegraphics[width=\textwidth]{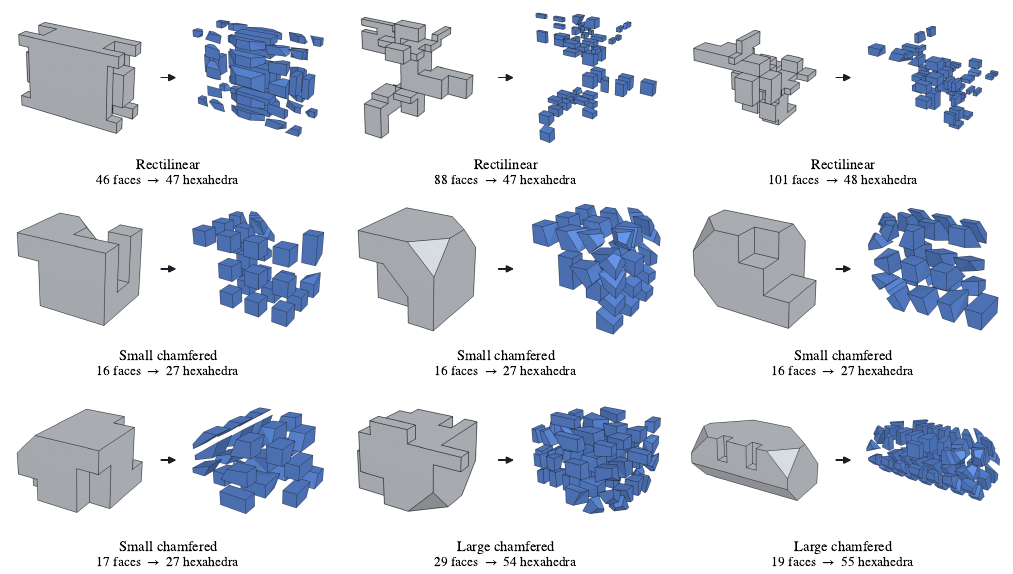}
\caption{Held-out test solids (grey) beside an accepted all-hexahedral
decomposition (exploded) by seed 1's $\pi_R$ at $\beta=3$ and quality threshold
$0.4$: three rectilinear, four small chamfered and two large chamfered solids,
each with its set, input face count and hexahedron count.}
\label{fig:gallery}
\end{figure*}

On held-out small chamfered solids whose boundary topology appears nowhere in
the cloning data, the final policy solves an expected $94.4\%$ of solids in
$20$ samples, and $99.4\%$ of held-out rectilinear solids
(\cref{fig:headline,fig:gallery}). Expert iteration adds most where the clone
is weakest, raising best-of-$20$ on large chamfered solids from $0.203$ to
$0.355$. A control with the same samples and updates on a frozen archive shows no
improvement, so the gains come from the new solutions that each
round adds. On rectilinear solids, $96\%$ of the decompositions reach a lower
bound on irregularity, and most have fewer cells than the grid decomposition.
On our three main test sets, Gmsh's all-hexahedral mode
\citep{geuzaine2009gmsh} meshes nearly every solid, but none of its meshes
meets the quality threshold, and they are an order of magnitude larger.

The policy-gradient methods we tried are not a good fit for this problem.
Starting from the same clone at matched rollouts and optimizer steps, PPO,
GRPO and VinePPO all make the policy worse on the validation set. POPO, an
on-policy update closer to expert iteration that trains only on won episodes,
improves the solids it trains on but not the validation set
(\cref{sec:exp:pg}).

Our contributions are:
\begin{itemize}
  \item a formulation of hexahedral block decomposition as a decision problem
        on 3-G-maps, with three inserts, geometry-free action masks, and a
        dart-convolution policy that makes one decision per inserted cell
        (\cref{sec:mdp,sec:policy});
  \item a training recipe that clones certified plans with set-valued targets
        and then improves the policy by expert iteration over a per-solid
        archive of its best decompositions (\cref{sec:training});
  \item an evaluation on held-out solids that are disjoint from the cloning
        data in boundary topology, with two training seeds, a no-growth
        control, a comparison with on-policy updates (PPO, GRPO, VinePPO and
        POPO) from the same clone at matched rollouts and optimizer
        steps, and an analysis of where the method fails and of the quality of its
        decompositions (\cref{sec:experiments}).
\end{itemize}

\section{Hexahedral block decomposition}
\label{sec:problem}

The input is a polyhedral solid $\Omega\subset\mathbb{R}^3$ given by its
boundary: planar polygonal faces meeting at \emph{ridges} (input edges) and
\emph{corners} (input vertices). A \emph{hexahedral block decomposition} of
$\Omega$ is a partition into cells, each combinatorially a cube (six
quadrilateral faces, twelve edges, eight vertices), glued face to face, whose
boundary faces tile $\partial\Omega$ and whose vertices on ridges and corners
stay on them. A decomposition is \emph{usable} when every cell is well shaped
after untangling; we measure shape by the minimum scaled Jacobian $\mathrm{SJ}$
over all cell corners ($1$ for a cube, $\le 0$ for an inverted cell).

\paragraph{The quality ceiling and the quality threshold.}
Some inputs cannot be meshed well at all: a sharp ridge or an obtuse corner
caps the scaled Jacobian of every cell touching it. We compute a per-input
\emph{ceiling} $c(\Omega)$, an upper bound on the minimum $\mathrm{SJ}$ of any
valid mesh whose input ridges and corners are at their target valences, from the dihedral angles of the ridges and the sector angles at the
corners (Appendix~\ref{app:floor}). A decomposition is accepted when it is
all-hexahedral and its untangled minimum scaled Jacobian reaches the quality
threshold
\begin{equation}
  g(\Omega) = \min\!\big(\bar g,\ 0.99\,c(\Omega)\big),
  \label{eq:gate}
\end{equation}
so a cube-like input must reach the cap $\bar g$ while an input whose geometry
forbids it is held to what is possible. The factor $0.99$ is a numerical
tolerance: a decomposition at the ceiling is not rejected for falling short of
it in the last digits after untangling. We use $\bar g=0.4$ for evaluation and
$\bar g=0.5$ for training, which are generally considered acceptable thresholds for engineering applications.

\paragraph{Irregularity.}
Every vertex and edge of a hex mesh has a target valence that a structured
grid would give it: six edges at an interior vertex, four faces at an interior
edge, and targets set by the local dihedral and sector angles on
$\partial\Omega$. Let $V$ and $E$ be the total absolute deviation of vertex
and edge valences from their targets. Among accepted decompositions the more
regular ones, with smaller $V+E$, are preferable. Irregularity is not part of
acceptance; we use it to report how regular the decompositions are
(Section~\ref{sec:exp:quality}), against a lower bound $\underline{I}(\Omega)$
on $V+E$, the \emph{floor} (Appendix~\ref{app:floor}).

\section{Decomposition as a sequential decision problem}
\label{sec:mdp}

\subsection{States: 3-generalised maps}
\label{sec:mdp:state}

\begin{figure}[t]
\centering
\includegraphics[width=\columnwidth]{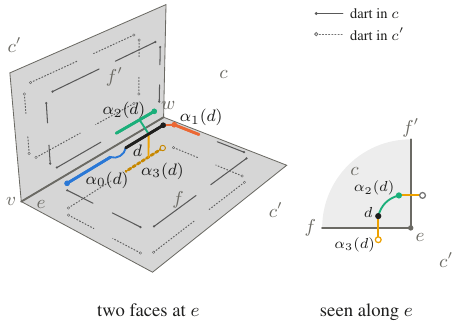}
\caption{The four involutions at one edge. A dart is a (vertex, edge, face,
volume) flag, drawn as a short stroke with a dot at its vertex (solid in the
volume $c$, dotted in neighboring volume $c'$); here $d=(w,e,f,c)$. Each $\alpha_i$ holds three of
the four defining elements of a dart constant and exchanges the $i$th element: $\alpha_0$ exchanges the vertex,
$\alpha_1$ the edge, $\alpha_2$ the face and $\alpha_3$ the volume.
\emph{Right:} the same edge viewed along its length. On a boundary face there
is no $c'$, and $\alpha_3(d)=d$.}
\label{fig:gmap}
\end{figure}

We represent the evolving decomposition as a 3-dimensional generalised map
(3-G-map; \citealp{damiand2014combinatorial}), the volumetric analogue of the
half-edge structure used in two dimensions \citep{narayanan2026playing}. A 3-G-map is a
tuple $M=(D,\alpha_0,\alpha_1,\alpha_2,\alpha_3)$ where $D$ is a finite set of
\emph{darts}, one per incident (vertex, edge, face, volume) flag, and each
$\alpha_i$ is an involution on $D$ that changes only the $i$-cell of the flag:
$\alpha_0$ swaps the vertex, $\alpha_1$ the edge, $\alpha_2$ the face and
$\alpha_3$ the volume (\cref{fig:gmap}). The involutions $\alpha_0,\alpha_1,\alpha_2$ have no
fixed points; a dart with $\alpha_3(d)=d$ lies on $\partial\Omega$. Cells are
not stored but recovered as orbits: the vertex of $d$ is its orbit under
$\langle\alpha_1,\alpha_2,\alpha_3\rangle$, its edge under
$\langle\alpha_0,\alpha_2,\alpha_3\rangle$, its face under
$\langle\alpha_0,\alpha_1,\alpha_3\rangle$ and its volume under
$\langle\alpha_0,\alpha_1,\alpha_2\rangle$. Vertex coordinates are stored
separately. Because every incidence is an involution, the representation is
uniform across dimensions and cell types, and the edits below are local
re-wirings of the $\alpha_i$.

An episode starts from $\Omega$ itself as a single volume whose
faces are the input polygons, and a state is the current G-map together with
its vertex coordinates and the move count.

\subsection{Actions: three inserts}
\label{sec:mdp:actions}

\begin{figure*}[t]
\centering
\includegraphics[width=\textwidth]{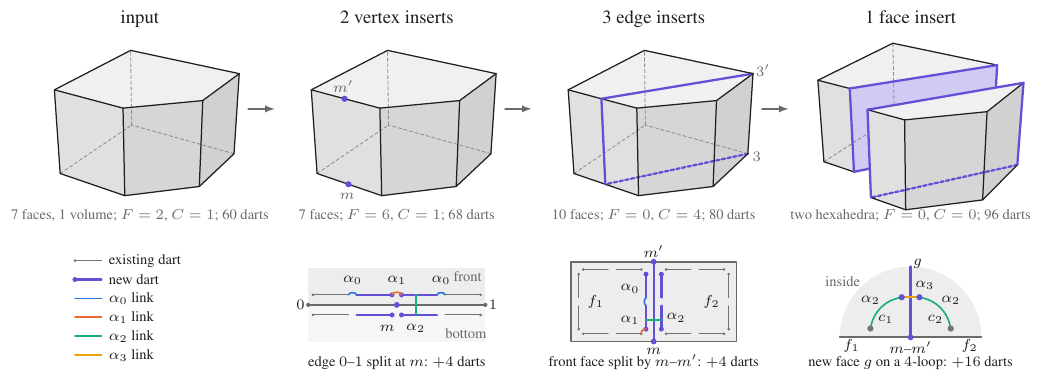}
\caption{One cut made of the three inserts. \emph{Top:} an edge-chamfered block (a
pentagonal prism, one volume) is split into two hexahedra. Two vertex inserts
split the front edges at $m$ and $m'$. Three edge inserts split the bottom,
top and front faces by $m$--$3$, $m'$--$3'$ and $m$--$m'$. These three new
edges and the existing edge $3$--$3'$ close a $4$-loop, and one face insert on
it splits the volume (the halves are pulled apart for display).
\emph{Bottom:} the darts each insert adds, drawn as in \cref{fig:gmap}. A
vertex insert adds one dart per dart of the split edge's orbit
$\langle\alpha_0,\alpha_2,\alpha_3\rangle$; an edge insert adds four darts
on each side of the face it splits;
a face insert adds four darts per edge of its loop.}
\label{fig:inserts}
\end{figure*}

The agent edits the map with three insertion operations
\citep[Ch.~4]{damiand2014combinatorial} (\cref{fig:inserts}):
\begin{itemize}
  \item a \textbf{vertex insert} splits an edge at a new vertex;
  \item an \textbf{edge insert} splits a face by a new edge between two of its
        vertices that are not adjacent along it;
  \item a \textbf{face insert} splits a volume by a new interior polygon whose
        boundary is a closed loop of $k\ge 3$ \emph{existing} edges of that
        volume.
\end{itemize}
The environment also supports the corresponding deletions. We do not offer
them to the agent, since they are not needed to reach the decompositions we
train on. Each expert plan is the reverse of a backward walk that dismantles a
constructed decomposition by deletions (Section~\ref{sec:training:certs}), and
every deletion is undone by one of the three inserts: a face deletion merges two
volumes, an edge deletion two faces, and a vertex deletion two edges. We call a group
of inserts that ends in one face insert a \emph{cut}.

\paragraph{Masks.}
Every action offered to the agent is valid. A vertex insert is offered once per
edge; an edge insert once per pair of non-adjacent vertices of a face, so that
neither side is a two-gon; a face insert is built step by step from a seed edge
by following edges of the volume, and only continuations from which the loop
can still close are offered (a bounded depth-first search over the volume's edge
graph; Appendix~\ref{app:impl}). No mask tests planarity,
alignment or any other geometric property of the input, so the same action set
applies to every polyhedron.

\paragraph{One decision per inserted cell.}
In the underlying environment an edge insert takes two picks (its two endpoints)
and a face insert of a $k$-loop takes $k$ picks. We expose instead one decision
per inserted cell: a vertex insert, a whole edge (an unordered vertex pair of a
face), or a face-insert seed followed by its loop
(Section~\ref{sec:policy:heads}).

\paragraph{Vertex placement.}
The agent chooses connectivity only. A new vertex starts at the midpoint of the
edge it splits. Each vertex stays on the part of the boundary it starts on:
input corners are fixed, vertices on a ridge slide along it, vertices on an
input face slide in its plane, and interior vertices are free. After every
insert we apply five Jacobi sweeps of Laplacian smoothing, each followed by
projection back onto the boundary. A stronger untangling pass runs only at
the acceptance test.

\subsection{Termination}
\label{sec:mdp:term}

Let $F=\sum_{f}|\mathrm{sides}(f)-4|$ and $C=\sum_{c}|\mathrm{faces}(c)-6|$
measure how far the faces and cells are from quadrilaterals and hexahedra, and
let $V$ and $E$ be the valence deviations of Section~\ref{sec:problem}.
An episode ends with success when every face is a quadrilateral and every cell
has six faces ($F=C=0$), every cell is a combinatorial cube, and the untangled
minimum scaled Jacobian reaches the quality threshold \eqref{eq:gate}. A mesh with $F=C=0$ that fails either
test is not terminal: a flag in the observation records the rejection and
the agent may keep editing. An episode is truncated when the raw defect
$V+E+F+C$ exceeds $8E_{\mathrm{in}}$ for an input with $E_{\mathrm{in}}$ edges
(we call this a \emph{defect blow-up}), or after $2T$ picks, where
$T=\max(64,\,8\,n_{\mathrm{out}})$ for $n_{\mathrm{out}}$ outward darts
(Section~\ref{sec:policy:trunk}) of the input. An episode with no legal move also ends, and every end other
than acceptance counts as a failure. Expert iteration uses no
reward; the policy-gradient arms of Section~\ref{sec:exp:pg} use a shaped,
potential-based reward (Appendix~\ref{app:reward}).

\subsection{Observation}
\label{sec:mdp:obs}

The mesh grows to thousands of darts, so the policy sees a crop window of at most
$N=512$ outward darts (Section~\ref{sec:policy:trunk}) around a \emph{centre}
cell. The window holds the darts of the centre cell first, then those of the
cells one face away, then two faces away, and so on, until it is full: a
breadth-first expansion over the dart neighbours $\sigma_1,\dots,\sigma_4$ of
Section~\ref{sec:policy:trunk}, in which only $\sigma_4$ crosses into another
cell. The centre
stays on its cell while that cell is not a hexahedron, then moves to the most
irregular cell in the last window, then to the most irregular cell in the mesh.
It is held fixed while a multi-pick insert is being built. A face-insert seed is
offered only in a cell the window holds whole, and an edge insert only on a face
the window holds whole.

Each dart in the window carries $23$ features: target and actual valences,
the sizes of its face and cell, its edge length relative to the window's
median, the corner and dihedral angles at its edge, boundary flags, the class
of its vertex relative to the input (corner, ridge, surface, interior) and
flags for an insert in progress (Appendix~\ref{app:impl}). No feature contains
a coordinate or a direction, so the observation is invariant to rotation,
translation and uniform scaling of $\Omega$. A global vector of ten scalars
(the defect terms relative to the initial defect, the mesh size, the worst
scaled Jacobian of the finished hexahedra and the rejection flag) and the
fraction of the budget used complete the observation. Every count, per dart
and global, is capped, so no feature grows without bound with the size of the
solid.

\section{A policy on darts}
\label{sec:policy}

\begin{figure*}[t]
\centering
\includegraphics[width=\textwidth]{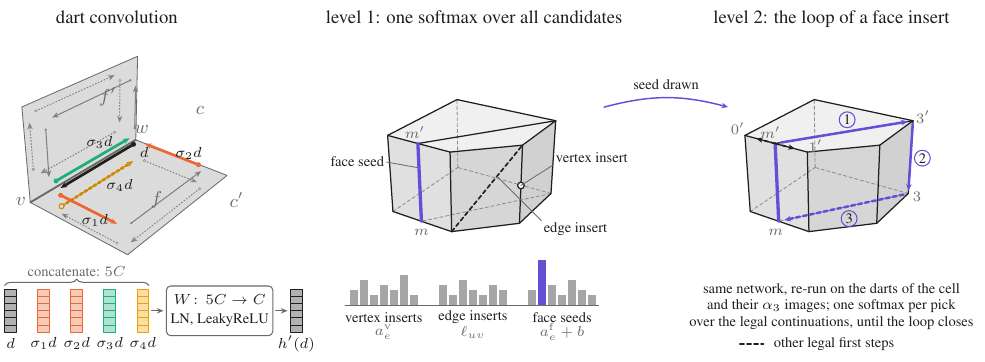}
\caption{The policy. \emph{Left:} the four neighbours $\sigma_1d,\dots,\sigma_4d$
of a dart $d$ in the dart convolution \eqref{eq:dartconv}, at the edge of
\cref{fig:gmap}. Outward darts are drawn as arrows from their vertex (solid in
$c$, dotted in $c'$), each $\sigma_k$ in the colour of the involution it
composes with $\alpha_0$. \emph{Middle:} one softmax ranks every vertex-insert,
edge-insert and face-seed candidate in the window, shown on the block of
\cref{fig:inserts} after its edge inserts (bar heights schematic).
\emph{Right:} after the seed $m$--$m'$ is drawn, the loop is decoded one edge
per pick; picking $3'$, then $3$, then closing at $m$ gives the face insert of
\cref{fig:inserts}.}
\label{fig:policy}
\end{figure*}

\subsection{Dart convolution}
\label{sec:policy:trunk}

Darts come in two orientation classes, a topological two-colouring in which
every involution $\alpha_i$ maps a dart to the other class. The policy only
ever sees the \emph{outward} class, which halves the input without losing
information, and so uses the four compositions of involutions that stay in it:
$\sigma_1=\alpha_1\alpha_0$ and $\sigma_2=\alpha_0\alpha_1$ (next and previous
dart around the face), $\sigma_3=\alpha_0\alpha_2$ (same edge, adjacent face of
the same volume) and $\sigma_4=\alpha_0\alpha_3$ (same face, adjacent volume).
These are the only neighbours a dart has (\cref{fig:policy}, left). One convolution block maps per-dart
features $h\in\mathbb{R}^{|D|\times C}$ to
\begin{equation}
  \begin{aligned}
  h'(d) = \mathrm{LeakyReLU}\Big(\mathrm{LN}\big(W\,[\,&h(d);\,h(\sigma_1 d);\,h(\sigma_2 d);\\
          &h(\sigma_3 d);\,h(\sigma_4 d)\,]\big)\Big),
  \end{aligned}
  \label{eq:dartconv}
\end{equation}
with $W\in\mathbb{R}^{C\times 5C}$, so the four relations keep separate weights.
A neighbour outside the crop window and a missing neighbour across
$\partial\Omega$ (where $\alpha_3(d)=d$) are replaced by two learned sentinel
vectors per block, so the network can tell ``not observed'' from ``boundary''.
This is the 3-G-map counterpart of the half-edge convolution of
\citet{narayanan2026playing}; like it, the operator has no notion of cell type, and its
receptive field grows along the map's own connectivity rather than in
Euclidean space. The trunk is a linear input projection followed by four
residual stages of two blocks each ($h\leftarrow h+B_2(B_1(h))$), with $C=96$
channels.

A context vector is computed once per state from the mean and maximum of the
final dart embeddings over the window together with the global vector of
Section~\ref{sec:mdp:obs}, by a two-layer MLP; it is added to every dart's latent
before the output heads. A linear head gives each dart one score per
elementary operation (vertex insert, edge-insert endpoint, face-insert seed and
loop step); the logits of the decisions below are built from these scores.
Each operation is offered on one \emph{canonical} dart per cell it acts on: the
first dart in the window's order of each edge (vertex insert), of each face
corner (edge-insert endpoint) and of each edge side within a volume (face
seed).

\subsection{Decisions}
\label{sec:policy:heads}

At each state the policy draws one decision from a single softmax over three
kinds of candidates (\cref{fig:policy}, middle); a face insert then continues
with the picks of its loop (\cref{fig:policy}, right).
\begin{itemize}
  \item \emph{Vertex insert} on edge $e$: the vertex-insert score of a canonical
        dart of $e$.
  \item \emph{Edge insert} between vertices $u$ and $v$ of a face: a joint pair
        score
        \begin{equation}
        \ell_{uv} = \mathrm{lse}(a_u,a_v)
          + \mathrm{MLP}([z_u{+}z_v, z_u{\odot} z_v, \phi_{uv}, \gamma_{uv}]),
        \label{eq:pair}
        \end{equation}
        where $\mathrm{lse}$ is log-sum-exp, $a_u$ is the endpoint score and $z_u$ the latent of the dart at
        $u$, and $\odot$ is the elementwise product. $\phi_{uv}$ holds
        combinatorial features of the split (the sizes of the face and of its
        two parts) and $\gamma_{uv}$ geometric features of $v$ seen from $u$
        (Appendix~\ref{app:impl}). We score the pair jointly, rather than choosing an
        anchor and then a second end, so that the far end is chosen with
        knowledge of the near end and the split it produces.
  \item \emph{Face insert} from seed edge $e$ in volume $c$: the face-seed score
        of $e$ plus one learned scalar shared by all seeds, which sets how
        often face inserts are drawn against the other two kinds. Once a seed is drawn, the loop is
        decoded pick by pick by the \emph{same} network re-run on a restricted
        observation (the darts of $c$ and their $\alpha_3$ images) with the
        in-loop and loop-head flags set; each pick is a softmax over the legal
        continuations. The loop decoder therefore has no parameters of its own
        and sees the volume being split at full resolution whatever its
        position in the crop window.
\end{itemize}
A training target is a whole inserted cell, so its log-probability is that of
its entry in the softmax plus, for a face insert, the log-probabilities of the
picks that build its loop. The same face can be built from several start edges
and directions; we store up to four of these pick sequences and train on
whichever the current policy finds most probable (hard EM). At test time all logits are multiplied
by an inverse temperature $\beta$ before every softmax.

The policy has about $0.46$M parameters.

\section{Training: from certificates to expert iteration}
\label{sec:training}

Success is known only at the end of an episode, and whether a partial
decomposition can still be completed is not visible until much later. A policy
that picks uniformly among legal moves never produces an accepted decomposition
of any test solid, chamfered or rectilinear: none of $6{,}000$ episodes on the
small chamfered test set and none of $12{,}000$ on the small-chamfered
validation set succeed (\cref{sec:exp:main}). We therefore train by imitation
only; \cref{sec:exp:pg} compares policy-gradient updates at matched compute.
The policy is cloned from demonstrations that are correct by construction, and
then improved by \emph{expert iteration}
\citep{anthony2017thinking}: sample the current policy on solids it does not
yet solve, keep only accepted decompositions, and clone those
(\cref{alg:ei}).

\subsection{Certified demonstrations}
\label{sec:training:certs}

For several families of solids an accepted decomposition is known in advance.
For a \emph{rectilinear} solid (a polycube, or a union and difference of
axis-aligned boxes), extending every input face plane through the domain and
merging redundant planes gives a grid decomposition. The same holds for a
polycube built on a grid whose planes are each tilted by up to $15^\circ$ in a
random direction. Star-shaped polygonal prisms have a
polar decomposition. One Catmull--Clark refinement of a convex polyhedron gives
one hexahedron per vertex. A \emph{swept prism} (a planar polygon extruded along
a straight path) is decomposed by extruding a quadrilateral mesh of its
cross-section into a single layer of hexahedra. A learned 2D quadrilateral
meshing policy \citep{narayanan2026playing} proposes the cross-sections, and
each extruded prism is kept only if its plan replays to an accepted hex mesh
that passes the quality threshold in our environment. Cross-sections that are
good enough in 2D but fall below the threshold after extrusion are rejected. For a small block whose convex
edges or corners are chamfered, we start from the grid certificate of the
un-chamfered block and replace each cell the chamfer cuts by a local template:
one Catmull--Clark refinement of the cut cell, or a dedicated corner template.
Each of these is an accepted decomposition by construction, a
\emph{certificate}. The clone $\pi_0$ is trained on certificates only: $2.61$M
states from $22{,}465$ solids, of which $2.00$M come from grid certificates,
$0.36$M from chamfer plans and $0.25$M from swept prisms
(\cref{tab:p0}, Appendix~\ref{app:lineage}). It is trained for six epochs, and
the epoch is chosen on validation. Each later round trains exactly one epoch,
with no selection.

A certificate is a target, not a demonstration: the agent needs the sequence of
inserts that builds it from the raw solid. We obtain it by a \emph{backward
walk}, deleting the certificate's interior faces one at a time and, after each
deletion, every edge and vertex the deletion frees (coplanar edge merges, then
collinear vertex merges, with input corners pinned), until only the input
solid remains. Read forwards, the walk is a plan in the agent's own action space,
and each group between face deletions is one cut. A plan is kept only if it
replays to an accepted state in the training environment, under the same masks
and budget as at test time and the training threshold ($\bar g=0.5$ in
\cref{eq:gate}). No
plan with a single failing move enters training.

\paragraph{The certificate gap.}
Certificates exist only where a decomposition can be written down. The solids we
care about here are polycube-like solids whose edges and corners have been
\emph{chamfered}: cut off by oblique planes, which produces triangular corner
faces and concave chamfer strips. A plane-extension grid of a chamfered solid
produces non-hexahedral cells at every chamfer. The template construction
certifies only part of the small chamfered solids, and on large chamfered solids
it mostly fails. Bridging this
gap is what expert iteration is for.

\subsection{Set-valued behaviour cloning}
\label{sec:training:bc}

Inserts on different parts of the mesh commute, so a plan is one of many
orderings of the same set of moves, and the plan's own order is an artefact of
the walk. We replay each plan in a uniformly random order and, at every visited
state $s$, record as target the set $\mathcal{T}(s)$ of \emph{all} remaining plan
moves that are playable in $s$. The loss is cross-entropy against the uniform
distribution over that set,
\begin{equation}
  \mathcal{L}_{\mathrm{BC}}(\theta) = -\,\mathbb{E}_{s}\Big[\tfrac{1}{|\mathcal{T}(s)|}
     \textstyle\sum_{m\in\mathcal{T}(s)} \log \pi_\theta(m\mid s)\Big],
  \label{eq:bc}
\end{equation}
where $\log\pi_\theta(m\mid s)$ is the decision log-probability of
Section~\ref{sec:policy:heads}, taken over the most probable of a face insert's
pick sequences. The set-valued target removes the plan's arbitrary order from
the supervision. Training on the single next move in plan order would give
conflicting labels: plans contain moves that commute, and on a symmetric solid,
darts related by the symmetry have identical inputs and hence identical
embeddings, so one order would label an embedding positive in one sample and
an identical one negative in another.

\subsection{Expert iteration}
\label{sec:training:ei}

\begin{algorithm}[t]
\caption{Expert iteration for hex decomposition}
\label{alg:ei}
\begin{algorithmic}
  \STATE {\bfseries Input:} certified plans $\mathcal{P}_0$, a fixed set
         $\mathcal{U}$ of training solids, none with a held-out signature
  \STATE $\pi_0 \leftarrow$ clone on $\mathcal{P}_0$ (\cref{eq:bc});
         archive $\mathcal{A}\leftarrow\emptyset$
  \FOR[at most $R=5$ rounds]{$i=1,\dots,R$}
    \FOR{each $\Omega\in\mathcal{U}$ with $\mathcal{A}[\Omega]=\emptyset$, and a
         random $10\%$ of the others}
      \STATE run restart search with $\pi_{i-1}$ at inverse temperature $\beta$
             until the move budget $B$ is spent
      \STATE insert each accepted episode into $\mathcal{A}[\Omega]$; keep it
             only if it ranks among the best (\cref{eq:rank})
    \ENDFOR
    \STATE $\pi_i\leftarrow$ fine-tune $\pi_{i-1}$ for one epoch on
           $\mathcal{P}_0\cup\mathcal{A}$
    \STATE stop if $\pi_{i-1}$ and $\pi_i$ both fail to improve the best
           validation score so far
  \ENDFOR
  \STATE \textbf{return} the round $\pi_r$, $0\le r\le i$, with the best
         validation score
\end{algorithmic}
\end{algorithm}

\paragraph{Mining.}
Mining runs on one fixed set $\mathcal{U}$ of $12{,}000$ training solids from
the same generators as the test solids: $3{,}000$ in each of four classes
(small chamfered, concave chamfered, large chamfered and rectilinear), defined
by geometry and generator size only. $\mathcal{U}$ was drawn once before the
first round, is fixed for both seeds and every round, and shares no signature
with a validation or test solid (below). Each round runs the current policy on the solids of
$\mathcal{U}$ it has not yet solved, plus a random $10\%$ of the solved ones,
on which a decomposition with fewer hexahedra may still be found. For each (solid, seed) we run a \emph{restart search}: sample
an episode at inverse temperature $\beta=3$; if it ends without success
(\cref{sec:mdp:term}), restart from the raw solid; and stop at the first accepted decomposition or when $6{,}000$
committed moves are spent (Appendix~\ref{app:search}).
Restart search is a mining tool only: all test numbers use plain independent
sampling.

\paragraph{The solution archive.}
Wins are not simply accumulated. For every training solid we keep an
\emph{archive} $\mathcal{A}[\Omega]$ of its best accepted decompositions,
ranked lexicographically: fewest hexahedra first, then the largest
$\mathrm{SJ}_{\min}$, then the fewest moves,
\begin{equation}
  \min\ \big(\,\#\text{hexahedra},\ \ -\mathrm{SJ}_{\min},\ \
  \#\text{moves}\,\big),
  \label{eq:rank}
\end{equation}
where $\mathrm{SJ}_{\min}$ is the minimum corner scaled Jacobian of the final
hexahedra as the environment placed them (before untangling). A second solution
of the same solid is kept only if it differs from the first in hexahedron count
or vertex count; at most two are kept. The archive therefore grows
only when mining solves a new solid or finds a better decomposition of a solved
one, and repeated wins on easy solids do not crowd out rare ones. An archived
decomposition is treated exactly like a certificate: it is replayed once in a
random order through the set-valued collector of Section~\ref{sec:training:bc},
with targets drawn from its own moves.

\paragraph{Training.}
Each round fine-tunes the previous checkpoint for one epoch on the
re-collected archive together with the certified plans $\mathcal{P}_0$,
which stay in every round as a fixed base group at a weight fixed before the
first round ($25\%$ of effective samples); the optimiser settings are in
Appendix~\ref{app:impl}. Every round starts from its predecessor, whatever its score: no round is
rejected or re-run. After each round we compute one validation score $V_r$,
the pooled $\mathbb{E}[\mathrm{b20}]$ over the small- and large-chamfered
selection sets (\cref{tab:sets})
at the round's reporting $\beta$ (\cref{sec:exp:protocol}), with $\pi_0$ as
round $0$. A round improves if $V_r$ exceeds the best earlier score by more
than $0.005$. The run stops after two consecutive rounds that do not improve,
judged per seed, and otherwise after $R=5$ rounds. The headline policy $\pi_R$ is
the round with the best $V_r$, which need not be the last one. The clean
small-chamfered validation set is never used for these choices. The rule was fixed before any expert-iteration round's
validation score was seen. Mining and training use the stricter threshold $\bar g=0.5$; the stopping
rule, the choice of $\pi_R$ and every reported number use $\bar g=0.4$.

\paragraph{Held-out hygiene.}
Generated solids repeat across random seeds, so we keep the training data
apart from the held-out sets in three steps. First, every solid has a shape
signature: its face count and its vertex positions, normalised for
translation, uniform scale and the $48$ symmetries of the cube. The certified
data are collected with a filter that removes every solid whose signature
appears in a validation or test set, so no exact copy of a held-out solid is
cloned. Second, $\mathcal{U}$ is drawn with the same filter, and a scan of
$\mathcal{U}$ and of every archived solid finds no validation or test
signature. Third, a stretched or re-spaced copy of a solid has a different
signature but the same boundary topology, so a plan for one replays on the
other. We therefore also compare boundary topology alone: the
incidence graph of the solid's vertices, edges and faces, with each face
labelled by its number of sides. We report a test solid only if no solid in the
cloning data $\mathcal{P}_0$ shares its boundary topology (\cref{tab:sets},
Appendix~\ref{app:neardup}). Matches with the expert-iteration archives are
kept, since the policy found those decompositions itself.

\subsection{Inference}
\label{sec:training:inference}

At test time we draw $K$ independent episodes at the round's reporting
inverse temperature (Section~\ref{sec:exp:protocol}) and report a solid as
solved if any of them is accepted. This is best-of-$K$; we use $K=20$ unless
stated otherwise.

\section{Experiments}
\label{sec:experiments}

We ask four questions. Does the method decompose held-out solids that have no
certificate (\cref{sec:exp:main})? How much does each round of expert iteration
contribute (\cref{sec:exp:ei}), and would a policy-gradient update do better
(\cref{sec:exp:pg})? Where does the method still fail
(\cref{sec:exp:failures})? And how good are its decompositions
(\cref{sec:exp:quality})?

\subsection{Setup}
\label{sec:exp:setup}

\paragraph{Solids.}
\label{sec:exp:sets}
All solids except the unseen corner types come from one generator: box unions
and differences on a lattice
(polycube-like solids with notches, steps and through-cuts), optionally with
edges and corners chamfered by oblique planes. A lattice \emph{ratio} sets the
size. \Cref{tab:sets} lists the evaluation sets. Every test set was drawn
before the expert-iteration rounds began, is used for reporting only, and was
never used to choose training solids, a checkpoint or a hyperparameter. We
report only solids whose boundary topology matches no solid in the cloning
data (\cref{sec:training:ei}). \emph{Rectilinear} holds $260$ large solids without
chamfers ($17$--$60$ grid cells), some of them tilted. \emph{Small chamfered},
the main test set, holds $100$ small chamfered solids. \emph{Large chamfered}
probes size. \emph{Unseen corner types} probes local structure absent from
training (out of distribution): a square, a pentagonal and a hexagonal pyramid
and an octahedron.

\begin{table}[t]
\caption{Evaluation sets. F: input faces. Test sets hold only solids whose
boundary topology matches no solid in the cloning data
(\cref{sec:training:ei}). \emph{Selection}: used to choose checkpoints
and $\beta$. The small-chamfered validation set is used for no choice.}
\label{tab:sets}
\centering
\footnotesize
\setlength{\tabcolsep}{3pt}
\resizebox{\columnwidth}{!}{%
\begin{tabular}{@{}llrl@{}}
\toprule
Set & Solids & $n$ & Role \\
\midrule
Rectilinear                & polycube / tilted, 17--60 cells & 260 & test \\
Small chamfered            & chamfered, ratio 4              & 100 & main test \\
Large chamfered            & chamfered, ratio 10, F $\ge$16  & 193 & test (size) \\
Unseen corner types        & pyramids, octahedron            & 4   & test (OOD) \\
\midrule
Small-chamfered selection  & chamfered, ratio 4              & 114 & validation (selection) \\
Large-chamfered selection  & chamfered, ratio 10             & 169 & validation (selection) \\
Rectilinear selection      & rectilinear                     & 72  & validation \\
Small-chamfered validation & chamfered, ratio 4              & 200 & validation (no choice) \\
\bottomrule
\end{tabular}}
\end{table}

\paragraph{Protocol.}
\label{sec:exp:protocol}
A solid is \emph{solved} when an episode ends accepted: all-hexahedral with
untangled minimum scaled Jacobian at the quality threshold \eqref{eq:gate}. We report the
per-sample solve rate (ps) and best-of-20 (b20). A single block of $20$
samples gives a noisy b20 (Appendix~\ref{app:impl}), so with $60$ samples per
solid we report the unbiased estimate
$\mathbb{E}[\mathrm{b20}] = \frac{1}{|S|}\sum_{\Omega\in S}\big(1-\binom{N-W_\Omega}{20}/\binom{N}{20}\big)$
for $W_\Omega$ wins out of $N=60$ \citep{chen2021evaluating}. Comparisons between
checkpoints are paired over solids, with $95\%$ intervals from $5{,}000$ level
bootstrap resamples. Every episode that does not end accepted is a failure.
Over all test-set evaluations, $99.6\%$ of lost episodes end by defect
blow-up; the rest reach the pick budget, have no legal move or end on a
sampling error. Apart from the clone's epoch (Section~\ref{sec:training:certs}),
the one validation-based selection in the pipeline is the reporting inverse
temperature. For each checkpoint, $\pi_0$ included, we
evaluate $\beta\in\{2,3,4\}$ and keep the value with the best
$\mathbb{E}[\mathrm{b20}]$ pooled over the small- and large-chamfered
selection sets, with near-ties going to $\beta=3$ (Appendix~\ref{app:impl}).
This choice affects only the reported
numbers, never the training data, which are always mined at $\beta=3$.

\paragraph{Checkpoints.}
We train the full recipe twice, with independent seeds. Each run is the
certificate clone $\pi_0$ followed by at most $R=5$ expert-iteration rounds,
with the validation stopping rule of \cref{sec:training:ei}. The headline
policy $\pi_R$ is the best round by the validation score $V_r$ ($\pi_0$
included). Best-of-$K$ is not
monotone across rounds on every set, so we also show every round of both
seeds (\cref{fig:rounds}).
Every result in this section is from these two runs, at the evaluation
threshold $\bar g=0.4$.

\paragraph{Baselines.}
We compare against two references that learn nothing. \emph{Uniform random}
runs in our environment with the same quality threshold, masks and budget, and draws every
decision uniformly over the legal candidates: the top-level move, the other end
of an edge insert, and each step of a face loop. It is our sampler at inverse
temperature $\beta=0$, with $60$ samples per solid. \emph{Gmsh}
\citep{geuzaine2009gmsh} in its all-hexahedral mode recombines elements where
it can and then applies its all-hexahedral subdivision, which splits each
tetrahedron into four hexahedra. We run it at element sizes $d$, $d/2$ and $d/4$
for the bounding-box diagonal $d$, and count a solid solved if any of the three
meshes is all-hexahedral and reaches the same per-solid threshold \eqref{eq:gate},
with the minimum scaled Jacobian taken as the better of the mesh as produced and
after smoothing its interior nodes. Our environment untangles with inserted
boundary vertices free to slide along the boundary, while Gmsh's boundary nodes
stay fixed, which handicaps Gmsh slightly. For both we also record the hexahedron
count and the irregularity excess $\max(0,V+E-\underline{I})$ under the same
targets and floor as our policy's decompositions.

\paragraph{Compute.}
Collecting the certified data takes $401$ CPU core-hours,
shared by both seeds. Training the clone takes $15.7$ GPU-hours on one A5000.
Each expert-iteration round then takes $1.8$--$2.7$ GPU-hours of training and
$1{,}880$--$2{,}870$ CPU core-hours, $70$--$82\%$ of it mining
(\cref{tab:compute}). A full run of five rounds takes $27.1$ GPU-hours and $11{,}834$ CPU core-hours (mean of the two seeds).

\subsection{Main results}
\label{sec:exp:main}

\begin{table*}[t]
\caption{Best-of-20 solve rate on the test sets (per-sample solve rate in
parentheses), quality threshold $\bar g=0.4$. Entries are the unbiased
$\mathbb{E}[\mathrm{b20}]$ from $60$ samples per solid. Unseen corners: solids
solved at least once in $60$ samples, out of $4$. Bold: best in each column
among the learned policies, for best-of-20 and per-sample separately. Seed
means: per-solid means over the two seeds, with $95\%$ intervals on the small
chamfered set. Paired change: $\pi_R-\pi_0$, seed means, paired over solids, with
$95\%$ intervals, for best-of-20 and per-sample rate. Learned
policies use the reporting $\beta$ chosen on validation; the clone $\pi_0$ is
trained on certificates only, and $\pi_R$ is the round with the best validation
score for each seed. Gmsh is deterministic at a given element size: its entry
is the fraction of solids for which its mesh at one of three element sizes is
all-hexahedral and passes the threshold.}
\label{tab:main}
\centering
\small
\resizebox{\textwidth}{!}{%
\begin{tabular}{@{}lccccc@{}}
\toprule
Policy & Rectilinear & Small cham. & Large cham. & Unseen corners \\
\midrule
\multicolumn{5}{@{}l}{\emph{Learned}} \\
Clone $\pi_0$, seed 1, $\beta{=}4$ & 0.991 (\textbf{0.837}) & 0.904 (0.430) & 0.187 (0.021) & 0/4 (0.000) \\
Clone $\pi_0$, seed 2, $\beta{=}4$ & 0.994 (0.836) & 0.926 (0.454) & 0.218 (0.033) & 0/4 (0.000) \\
$\pi_R$ seed 1 (round 5, $\beta{=}3$) & \textbf{0.996} (0.829) & \textbf{0.950} (\textbf{0.537}) & \textbf{0.383} (\textbf{0.072}) & 0/4 (0.000) \\
$\pi_R$ seed 2 (round 3, $\beta{=}3$) & 0.992 (0.789) & 0.939 (0.521) & 0.328 (0.057) & 0/4 (0.000) \\
\midrule
\multicolumn{5}{@{}l}{\emph{Seed means, best-of-20}} \\
Clone $\pi_0$ & 0.992 & 0.915 [0.870, 0.953] & 0.203 & 0/4 \\
$\pi_R$ & 0.994 & 0.944 [0.907, 0.976] & 0.355 & 0/4 \\
\midrule
\multicolumn{5}{@{}l}{\emph{Paired change, $\pi_R-\pi_0$}} \\
Best-of-20 & $+0.001$ [$-0.002$, $+0.005$] & $+0.030$ [$+0.011$, $+0.052$] & $+0.153$ [$+0.117$, $+0.189$] & --- \\
Per-sample & $-0.028$ [$-0.038$, $-0.017$] & $+0.087$ [$+0.058$, $+0.117$] & $+0.038$ [$+0.027$, $+0.049$] & --- \\
\midrule
\multicolumn{5}{@{}l}{\emph{Non-learned references}} \\
Uniform random, $\beta{=}0$ & 0.000 (0.000) & 0.000 (0.000) & 0.000 (0.000) & 0/4 (0.000) \\
Gmsh hex mode (best of 3 sizes) & 0.000 & 0.000 & 0.000 & 2/4 \\
\bottomrule
\end{tabular}}
\end{table*}

\Cref{tab:main} gives the headline. On the main test set, small chamfered,
the final policy $\pi_R$ reaches $\mathbb{E}[\mathrm{b20}]=0.944$, with a
per-sample rate of $0.529$. Every lost episode of $\pi_R$ on this set ends by
defect blow-up. Across the test sets, the median solved solid takes $153$
decisions, and one in ten takes more than $296$; a won episode on the small
chamfered set takes $9.5$\,s on one CPU core.

Most of this comes from the clone. Trained on certified decompositions alone,
$\pi_0$ reaches $0.915$ on the small chamfered set and $0.992$ on the
rectilinear set. Expert iteration adds most where the clone is weakest, on the
large chamfered solids (\cref{sec:exp:ei}). No policy solves any of the solids
with unseen corner types, though the network can represent their decompositions
(\cref{sec:exp:failures}).

Neither reference comes close on the test sets. Uniform random completes no
decomposition on any test set or on the small-chamfered validation set. Gmsh
produces an all-hexahedral mesh of nearly every solid, but at no element size
does any of its meshes pass the quality threshold on a test set other than the
unseen-corner one (\cref{tab:coarse}).

\subsection{What expert iteration buys}
\label{sec:exp:ei}

\begin{figure}[t]
\centering
\includegraphics[width=\columnwidth]{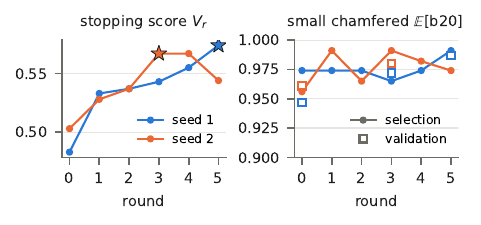}
\caption{Recipe rounds on validation. Left: the stopping score $V_r$, pooled
$\mathbb{E}[\mathrm{b20}]$ over the small- and large-chamfered selection sets at
each round's reporting $\beta$; stars mark $\pi_R$. Right:
$\mathbb{E}[\mathrm{b20}]$ on the small-chamfered selection set (lines, $20$
samples per solid) and on the small-chamfered validation set (squares, $60$
samples; scored for $\pi_0$, $\pi_R$ and seed 1's round 3), which plays no part
in any choice. \Cref{tab:recipe} (Appendix~\ref{app:recipe}) lists every
round.}
\label{fig:rounds}
\end{figure}

\Cref{fig:rounds} traces the rounds. The stopping score rises over the first
rounds for both seeds; seed 1 keeps improving to round 5, and seed 2 peaks at
round 3, so the rule picks those rounds. Mining finds fewer new solids each
round and replaces more archived solutions with better ones
(\cref{tab:recipe}). On validation, $\pi_R$ is ahead of $\pi_0$ in pooled
$\mathbb{E}[\mathrm{b20}]$ by $0.078$ on average ($+0.092$ $[+0.053,+0.133]$
and $+0.064$ $[+0.025,+0.104]$). On the small-chamfered
validation set, which plays no part in these choices, its per-sample rate is
$+0.084$ $[+0.068,+0.101]$ above $\pi_0$ and its $\mathbb{E}[\mathrm{b20}]$
$+0.030$ $[+0.015,+0.046]$ (seed means).

On the test sets, expert iteration changes most where the clone is weakest
(\cref{tab:main}, paired change). On the large chamfered solids best-of-20
rises from $0.203$ to $0.355$, and the solids never solved in $60$ samples
fall from $133$ to $103.5$ of $193$. The per-sample rate there rises too but
stays low ($0.065$), so on these solids search does most of the work. On the
small chamfered solids the per-sample rate rises by $0.087$ and best-of-20 by
$0.030$. On the rectilinear solids best-of-20 stays at its ceiling, while the
per-sample rate falls slightly.

Each round trains on an archive that has grown by that round's new wins, and
also trains for longer in total. To separate the two, a control retrains seed
1's rounds 2--5 from round 1's policy $\pi_1$ on the frozen round-1 archive,
with no further mining, matching each main round's samples and optimizer
updates. On validation the control's round 5 ends at $V=0.540$, where $\pi_1$
is at $0.533$ ($+0.007$ $[-0.028,+0.041]$), while the main
run's round 5 reaches $0.574$ ($+0.041$ $[+0.009,+0.074]$ above $\pi_1$).
Extra training on the same solutions does not move validation, so the main
run's gain comes from the new solutions that mining adds. The control was run
on one seed, and its direct difference from the main run's round 5
($-0.034$ $[-0.072,+0.005]$) just includes zero.

\subsection{On-policy updates}
\label{sec:exp:pg}

\begin{figure}[t]
\centering
\includegraphics[width=\columnwidth]{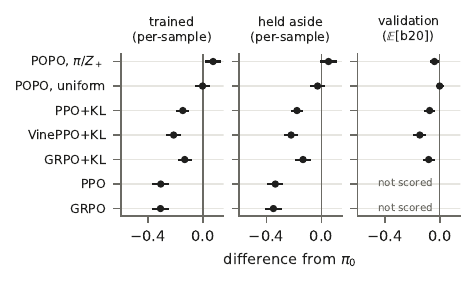}
\caption{Update rules at matched rollouts and optimizer steps, from seed 1's
$\pi_0$, after three iterations on $64$ small chamfered solids: difference from
$\pi_0$ with $95\%$ paired intervals. Trained and held aside: per-sample solve
rate on the $64$ trained and the $64$ held-aside solids ($16$ samples each).
Validation: $\mathbb{E}[\mathrm{b20}]$ on the small-chamfered validation set
($40$ samples per solid). Threshold $\bar g=0.4$, $\beta=3$. ``$+$KL'': with
the KL guard. \Cref{tab:pg} (Appendix~\ref{app:pg}) gives the values and the
on-policy wins.}
\label{fig:pg}
\end{figure}

Expert iteration improves the policy by cloning its own accepted episodes.
We ask whether a policy-gradient update on the policy's own episodes can do
the same. We compare several such updates at matched rollouts and optimizer
steps, on a small scale, each against the policy it starts from. Every arm
starts from seed 1's $\pi_0$. The
pool is $128$ small chamfered solids from the recipe's training set
$\mathcal{U}$ on which $\pi_0$'s round-1 restart search won between $10\%$ and
$70\%$ of its attempts; $64$ are trained on and $64$ are held aside. Each arm
runs three on-policy iterations. An iteration samples $16$ episodes per
training solid ($1{,}024$ episodes), then takes $1{,}000$ optimizer steps of $64$ moves with the
fine-tuning settings of \cref{sec:training:ei}. The first iteration of every
arm trains on the same episodes of $\pi_0$. The arms are:
\begin{itemize}
\item \emph{PPO} \citep{schulman2017proximal}, with generalised advantage
  estimates \citep{schulman2016high} from a learned critic.
\item \emph{GRPO} \citep{shao2024deepseekmath}: PPO's clipped objective with
  the episode return normalised within each solid's $16$ episodes, and no critic.
\item \emph{VinePPO} \citep{kazemnejad2024vineppo}: PPO with Monte Carlo
  advantages, from $4$ continuations at each of $11$ points along an episode.
\item \emph{POPO} \citep{xu2026popo}: an update on won episodes only,
  with an EMA anchor and a similarity penalty, weighting a solid's wins by
  $\pi(y)/Z_+$ or uniformly.
\end{itemize}
The policy-gradient arms train on the environment's shaped, potential-based
reward (Appendix~\ref{app:reward}); POPO uses no reward.
PPO and GRPO run unguarded and with a KL guard, which ends an iteration's
updates once the mean approximate KL over the last $20$ steps exceeds $0.05$;
VinePPO runs with the guard. All settings come from an earlier pilot, and
none were tuned here. We score the final policies on the pool and on the
small-chamfered validation set (\cref{fig:pg}).

\Cref{fig:pg} gives the result. One update improves the solids it trains on:
POPO with $\pi/Z_+$ weights, by $0.074$ per-sample over $\pi_0$; on the
held-aside solids its gain is not significant. With uniform weights POPO does
not move. Every policy-gradient update makes the policy worse on both halves,
and unguarded PPO and GRPO collapse (\cref{tab:pg}). The guarded arms stop at step $20$ of $1{,}000$ in every
iteration. Step $20$ is the first point at which the guard can fire, and the
mean approximate KL there is already $0.091$--$0.192$, against the $0.05$
target. So each iteration makes only $20$ updates, and those $20$ still lose a
third to more than half of the on-policy wins by iteration 3.

On the small-chamfered validation set, which neither the pool nor the
training touches, no arm beats $\pi_0$. Uniform POPO keeps $\pi_0$'s coverage
without improving it. POPO with $\pi/Z_+$ weights, which improved the trained
solids, loses coverage ($-0.039$ in $\mathbb{E}[\mathrm{b20}]$) while its
per-sample rate does not change significantly. Every policy-gradient arm
loses more: guarded PPO and GRPO least ($-0.074$ and $-0.081$), VinePPO most
($-0.146$).

Earlier in development we tried PPO extensively, with earlier cloned policies:
without a per-update KL limit it collapsed (held-out solve rate from $0.35$ to
$0.00$ within $8{,}000$ episodes), and with one it stayed near the cloned
policy. We therefore compare update rules at matched compute on a small pool.

\subsection{Where the method fails}
\label{sec:exp:failures}

\paragraph{Which solids fail.}
Concave chamfers are the hardest single feature. On the small-chamfered
selection set $\pi_R$'s best-of-20 misses one solid for each seed, in both
cases one with a concave chamfer; the $12$ of its solids with a concave chamfer have a
per-sample rate of $0.32$--$0.33$, against $0.74$--$0.75$ for the $102$ without.
On the large-chamfered selection set ($10$ samples each) the $82$ with a
concave chamfer are solved within ten samples $10\%$ of the time, against
$48$--$51\%$ for the $87$ without, and they are $62$--$63\%$ of the solids
$\pi_R$ does not solve. Corner chamfers matter less: on that set the solved
share is $0.31$--$0.36$ with up to two of them and $0.20$--$0.24$ with three or
more.

\begin{figure}[t]
\centering
\includegraphics[width=\columnwidth]{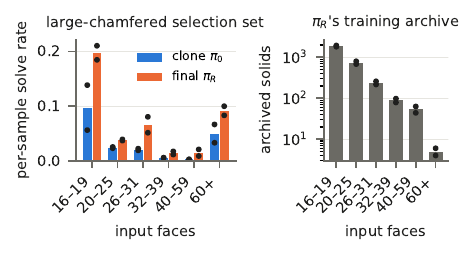}
\caption{Large-chamfered selection set ($169$ solids, $10$
samples each). Left: per-sample solve rate by number of input faces, for the
clone $\pi_0$ and the final policy $\pi_R$; bars are the mean of the two seeds,
dots the seeds. Right: chamfered solids in the archive $\pi_R$ was trained on,
in the same bins (log scale).}
\label{fig:size}
\end{figure}

\paragraph{Size.}
On the large-chamfered selection set the final policy's per-sample
solve rate is highest at $16$--$19$ input faces ($0.21$ and $0.18$ for the two
seeds, against $0.06$ and $0.14$ for $\pi_0$) and drops to $0.04$ at $20$--$25$
and $0.01$--$0.08$ for $26$--$59$ faces (\cref{fig:size}). The six solids with
$60$ or more faces are solved about as often ($0.08$--$0.10$). The training
archive thins out with size in the same way (\cref{fig:size}, right). This is consistent with the limit being the reach of mining at this
size rather than the architecture; above $40$ faces the $512$-dart window also no longer holds the cell
being cut.

\paragraph{Out of distribution.}
No recipe policy, $\pi_0$ or $\pi_R$, solves any of the four solids with
unseen corner types in $60$ samples. All $960$ of their episodes end by defect
blow-up. Each solid has a convex corner where four or
more faces meet, a corner type no training solid has, and none matches a
training solid even in boundary topology (\cref{app:neardup}). Each has a
decomposition that starts with a face insert on a plane that extends no input
face (through the apex); no training solid needs such a cut. Gmsh is the stronger method here: at its coarsest
element size it passes the threshold on the square pyramid ($8$ hexahedra) and
the octahedron ($16$), neither at the irregularity floor, and it fails on the
pentagonal and hexagonal pyramids (best minimum scaled Jacobian $0.28$ and
$0.27$). Some all-hexahedral mesh of these solids is easy to find: cut the
solid into tetrahedra and split each into four hexahedra, which is how Gmsh's
all-hexahedral mode works. A coarse, well-shaped decomposition is not. A
convex corner where four or more faces meet cannot be the corner of a single
hexahedron, so several cells must share the apex and split its solid angle.
A related problem in which the quadrilateral boundary
mesh is fixed in advance, Schneiders' pyramid, has a line of work of its own
\citep{yamakawa2010schneiders,verhetsel2019pyramid}; here the boundary mesh is
left free.

As a capacity check on the test solids themselves, not a generalisation
result, we built decompositions of these four solids for this test only (each
solid cut into tetrahedra, each tetrahedron split into four hexahedra) and
trained seed 1's $\pi_R$ for $10$ more epochs on them alone: the
decompositions in the order the construction makes its cuts, and random-order
replays of them. Sampled at $\beta=3$
without the defect blow-up limit, it then decomposes every one of the four into an
all-hexahedral mesh, in $59$, $49$, $9$ and $13$ of $60$ samples (square,
pentagonal and hexagonal pyramid, octahedron). Its meshes reproduce the
constructed decompositions, with the same number of hexahedra ($8$, $12$, $16$,
$16$) and the same minimum scaled Jacobian ($0.41$, $0.34$, $0.22$, $0.50$);
on the pentagonal and hexagonal pyramids these are themselves below our
quality threshold. The constructions are not necessarily the best
decompositions of these shapes: on the hexagonal pyramid, Gmsh's best mesh
($0.27$) is better shaped. Trained on the construction-order decompositions
alone, the policy decomposes only the square and pentagonal pyramids and the
hexagonal pyramid once: with many commuting moves on these symmetric solids,
it needs varied move orders. The network and move set can therefore represent
these decompositions. We expect that expert iteration fails here because no
training input has these corner types; adding constructed decompositions of
such corners to the cloning data is future work.

\subsection{Quality of the decompositions}
\label{sec:exp:quality}

An accepted decomposition is all-hexahedral by definition. The question is
whether it is a good \emph{block} decomposition: few cells, regular, and well
shaped. On the rectilinear test set we can compare against a reference,
the plane-expansion certificate (every input face plane extended through the
solid, redundant planes merged). Tilted
solids, which have no axis-aligned grid, are left out, so $218$ of the $260$
solids remain. For each solved solid we take the policy's decomposition with
the fewest hexahedra among $20$ samples.

\begin{table}[t]
\caption{The policy's decompositions against the plane-expansion certificate on
the $218$ untilted rectilinear test solids, $20$ samples per solid, for the clone $\pi_0$ and the final policy $\pi_R$ of each training
seed. Ratio: the policy's fewest-hexahedron count over the certificate's
hexahedron count, as a median with interquartile range. C\,/\,E\,/\,F: solids
where the policy is coarser, equal or finer. Hex.: median hexahedra,
certificate $\to$ policy. Floor: share of solved solids whose irregularity
equals the floor $\underline{I}$. SJ: median untangled minimum scaled Jacobian
(threshold $0.4$).}
\label{tab:quality}
\centering
\footnotesize
\setlength{\tabcolsep}{3pt}
\resizebox{\columnwidth}{!}{%
\begin{tabular}{@{}llcccccc@{}}
\toprule
Set & Policy & Solved & C\,/\,E\,/\,F & Ratio & Hex. & Floor & SJ \\
\midrule
Rectilinear & $\pi_0$ s1 & 217/218 & 142/72/3 & 0.96 [0.93, 1.00] & 37 $\to$ 34 & 98.6\% & 0.86 \\
            & $\pi_0$ s2 & 216/218 & 142/71/3 & 0.96 [0.93, 1.00] & 37 $\to$ 34 & 98.1\% & 0.85 \\
            & $\pi_R$ s1 & 217/218 & 145/68/4 & 0.96 [0.92, 1.00] & 37 $\to$ 34 & 96.8\% & 0.86 \\
            & $\pi_R$ s2 & 214/218 & 146/66/2 & 0.96 [0.91, 1.00] & 37 $\to$ 33 & 95.8\% & 0.86 \\
\bottomrule
\end{tabular}}
\end{table}

$\pi_R$ is coarser than the certificate on two thirds of the rectilinear
solids it solves and finer on almost none (\cref{tab:quality}). At the same
time, $96\%$ of these decompositions reach the irregularity floor exactly, and
their median minimum scaled Jacobian is $0.86$, well above the threshold. No axis or plane structure enters
the policy's input, and the policy does not reproduce the grid: it finds
decompositions that are as regular as the grid and smaller. This is already
true of the clone, which is coarser than the certificates it was trained on
for $66\%$ of the solved solids. Expert iteration changes the decompositions
little, and makes them slightly less regular: on $90\%$ of the solids both
solve, $\pi_R$'s fewest-hexahedron decomposition has the same count as
$\pi_0$'s, and the at-floor share drops by $2$ points.

On chamfered solids there is no plane-expansion certificate to compare
against (\cref{tab:coarse}). On the small chamfered set only $4\%$ of $\pi_R$'s
decompositions reach the irregularity floor. Because the floor is a bound that
need not be attained (Appendix~\ref{app:floor}), the median excess of $18$ is
an upper bound on the true excess. Chamfer corners cap the quality
ceiling of most of these solids at $0.5$ ($69$ of the $98$ whose ceiling is
known), and every accepted decomposition has minimum scaled Jacobian above
$0.4$. Against $\pi_0$ on the solids both solve, the hexahedron count is
unchanged on $65\%$ of them. On the large chamfered solids $\pi_R$ solves
$89.5$ of $193$ in $60$ samples ($\pi_0$: $60$), and one decomposition reaches
the floor. The $36$ solids that only $\pi_R$ solves
get larger decompositions (median $59$), while on solids both solve $\pi_R$'s
decomposition is smaller by $2.5$ hexahedra on average (equal on $38\%$).

\begin{table}[t]
\caption{$\pi_R$ against Gmsh on the test sets. $\pi_R$: median hexahedra and
median irregularity excess $V+E-\underline{I}$ of its fewest-hexahedron
decomposition in $60$ samples, on the solids it solves (seed means). Gmsh: the
same at its coarsest all-hexahedral element size; it is all-hexahedral on
$246$ of $260$, $100$ of $100$ and $192$ of $193$ solids. Its excess is
computed only for meshes under $2{,}000$ hexahedra ($243$ of $260$ and $179$ of
$193$ solids), so these medians understate it. Ratio: median over solids that
both mesh of $\pi_R$'s count over Gmsh's smallest mesh (largest $0.21$). SJ:
Gmsh's median minimum scaled Jacobian.}
\label{tab:coarse}
\centering
\footnotesize
\setlength{\tabcolsep}{3pt}
\resizebox{\columnwidth}{!}{%
\begin{tabular}{@{}lrrrrrr@{}}
\toprule
 & \multicolumn{2}{c}{Hexahedra} & & \multicolumn{2}{c}{Excess} & \\
\cmidrule(lr){2-3}\cmidrule(lr){5-6}
Set & $\pi_R$ & Gmsh & Ratio & $\pi_R$ & Gmsh & Gmsh SJ \\
\midrule
Rectilinear     & 32.5 & 878 & 0.03 & 0    & 3{,}184 & 0.05 \\
Small chamfered & 21   & 296 & 0.07 & 18   & 1{,}071 & 0.11 \\
Large chamfered & 49   & 756 & 0.07 & 31.5 & 2{,}594 & 0.06 \\
\bottomrule
\end{tabular}}
\end{table}

Gmsh's meshes are an order of magnitude larger (\cref{tab:coarse}): on every
solid that both mesh, $\pi_R$'s fewest-hexahedron decomposition is smaller
than Gmsh's smallest mesh, and no Gmsh mesh reaches the floor. This is expected of an
unstructured method, which tetrahedralises and then subdivides: it has no notion
of a coarse block layout. We include it as the most widely available automatic
all-hexahedral mesher, not as a competitor in block decomposition.

\section{Related work}\label{sec:related}

\paragraph{Hexahedral meshing and block decomposition.}
All-hexahedral meshing of general solids is still open
\citep{pietroni2022hex}: whether a quadrilateral boundary mesh admits a
compatible hexahedral mesh is a global question
\citep{mitchell1996characterization,eppstein1999linear}, and even small
inputs can call for extensive search \citep{verhetsel2019finding}. Practical
pipelines fix a structure first and then fill it: a polycube map
\citep{gregson2011allhex,livesu2013polycut}, an integer-grid map from a frame
field \citep{nieser2011cubecover,li2012allhex,liu2018singularity}, or an octree
refined to the boundary \citep{schneiders1996grid,marechal2009advances,gao2019feature};
Gmsh's all-hexahedral mode subdivides tetrahedra \citep{geuzaine2009gmsh}.
Multi-block structured meshing partitions a domain into blocks by hand-designed
geometric rules \citep{armstrong2015common}, and singularity index formulas
constrain where the blocks' singularities can go \citep{fogg2018singularities};
our floor applies identities of this kind. None of these methods learns a
policy over topological edits.

\paragraph{Learning to mesh.}
Reinforcement learning has been applied to quadrilateral mesh generation
\citep{pan2023reinforcement}, to improving the vertex regularity of triangle
and quadrilateral meshes by local and global element operations
\citep{narayanan2024learning}, to quadrilateral block decompositions of planar
domains that reach the irregularity floor set by the discrete Gauss--Bonnet
identity \citep{narayanan2026playing}, and to block decomposition of planar
CAD models \citep{diprete2024reinforcement}. In three dimensions,
\citet{zhang2025reinforcement} apply reinforcement learning to block
decomposition of solid models for hexahedral meshing, and
\citet{luo2026lightweight} decompose CAD models into sweepable blocks by
online reinforcement learning, without pre-training or offline data.
DL-Polycube \citep{yu2025dlpolycube} uses deep learning within the polycube
method rather than a sequence of edits; \citet{owen2026survey} survey the
area. Our policy acts on a 3-G-map of the decomposition itself, one policy
serves all solids, it is cloned from certified decompositions and improved by
expert iteration, and every accepted decomposition passes a quality threshold.

\paragraph{Expert iteration and filtered self-imitation.}
Expert iteration alternates search with imitation of its results
\citep{anthony2017thinking,silver2017mastering}; without a learned search it
becomes filtered self-imitation, as in self-imitation learning
\citep{oh2018self} and, for language models, STaR \citep{zelikman2022star},
rejection-sampling fine-tuning \citep{yuan2023scaling,dong2023raft} and ReST
\citep{gulcehre2023rest}, which \citet{havrilla2024teaching} find comparable
to or better than PPO for reasoning. Our method is closest to these, and
differs in three ways: it keeps an archive of the best decompositions found for
each solid, trains on set-valued targets that accept any order of a
decomposition's moves, and mines new decompositions by restart search over a
fixed set of training solids.

\paragraph{Policy gradients under sparse reward.}
PPO \citep{schulman2017proximal} with generalised advantage estimates
\citep{schulman2016high} is the standard on-policy method. GRPO replaces the
critic by a baseline from a group of samples \citep{shao2024deepseekmath},
VinePPO estimates values by Monte Carlo rollouts from intermediate states
\citep{kazemnejad2024vineppo}, and POPO trains on positive rollouts only, with
self-distillation \citep{xu2026popo}. Section~\ref{sec:exp:pg} compares these
updates at matched compute. None of these updates improved the cloned policy
on the validation set, and PPO, GRPO and VinePPO made it worse, likely because
success is a sparse
terminal event and the shaped reward (Appendix~\ref{app:reward}) is far from
monotone along a decomposition, which must make the mesh much less regular
before it can finish. The
problem may therefore serve as a benchmark for reinforcement learning on
sparse-reward, long-horizon tasks. GFlowNets train a sampler of compositional
objects in proportion to their reward \citep{bengio2021flow} and have been
applied to graph combinatorial optimization \citep{zhang2023let}; their
objectives need a backward policy over the states that lead to a given state,
which edit sequences on a mesh do not make easy to define.

\paragraph{Representations.}
Generalised maps represent the cells of a subdivided manifold as orbits of
involutions on darts \citep{lienhardt1994ndimensional,damiand2014combinatorial},
generalising the quad-edge structure of planar subdivisions
\citep{guibas1985primitives}. Edge and half-edge convolutions learn on fixed
triangle meshes \citep{hanocka2019meshcnn,ludwig2023halfedgecnn}; our dart
convolution runs on a map that the policy itself edits, and the map is the
state of the decision process.

\section{Conclusion}\label{sec:conclusion}

We posed hexahedral block decomposition as a sequence of topological inserts
on a 3-G-map and trained a single dart-convolution policy for it, by cloning
constructed decompositions and then by expert iteration over a per-solid
archive of its own best decompositions. The policy decomposes held-out
chamfered solids for which no construction exists, and expert iteration
helps most on the larger ones, through the new solutions each round adds
rather than through more training. At matched compute, policy-gradient
updates made the cloned policy worse.

The method has clear limits. Solids with concave chamfers are solved much less
often than those without. The solve rate falls with size, as the archive thins
out above about $20$ input faces, and above about $40$ faces the $512$-dart
window no longer holds the cell being cut. Solids with corner types that no
training solid has are not solved, although the network and move set can
represent their decompositions once they are shown. On chamfered solids few
decompositions reach the irregularity floor, which is itself only a bound in
three dimensions. The inputs are planar polyhedra without holes. The 3-G-map
represents holes as well: as in the planar case \citep{narayanan2026playing},
a \emph{slit} makes the input simply connected. A slit edge joins a face's
hole to its outer boundary and a slit face cuts a solid with a through-hole
into a ball, the face or volume meeting itself across the slit, and both are
valid 3-G-maps. Learning on such inputs needs demonstrations and inserts that
handle the slit. Natural next steps are holes, constructed decompositions for
new corner types in the cloning data, an observation that scales with the
mesh, and curved boundaries.

We will release our code.

\bibliography{refs}

\begin{thebibliography}{47}
\providecommand{\natexlab}[1]{#1}
\providecommand{\url}[1]{\texttt{#1}}
\expandafter\ifx\csname urlstyle\endcsname\relax
  \providecommand{\doi}[1]{doi: #1}\else
  \providecommand{\doi}{doi: \begingroup \urlstyle{rm}\Url}\fi

\bibitem[Anthony et~al.(2017)Anthony, Tian, and Barber]{anthony2017thinking}
Anthony, T., Tian, Z., and Barber, D.
\newblock Thinking fast and slow with deep learning and tree search.
\newblock In \emph{Advances in Neural Information Processing Systems}, 2017.

\bibitem[Armstrong et~al.(2015)Armstrong, Fogg, Tierney, and Robinson]{armstrong2015common}
Armstrong, C.~G., Fogg, H.~J., Tierney, C.~M., and Robinson, T.~T.
\newblock Common themes in multi-block structured quad/hex mesh generation.
\newblock \emph{Procedia Engineering}, 124:\penalty0 70--82, 2015.
\newblock \doi{10.1016/j.proeng.2015.10.123}.

\bibitem[Bengio et~al.(2021)Bengio, Jain, Korablyov, Precup, and Bengio]{bengio2021flow}
Bengio, E., Jain, M., Korablyov, M., Precup, D., and Bengio, Y.
\newblock Flow network based generative models for non-iterative diverse candidate generation.
\newblock In \emph{Advances in Neural Information Processing Systems}, 2021.

\bibitem[Chen et~al.(2021)Chen, Tworek, Jun, et~al.]{chen2021evaluating}
Chen, M., Tworek, J., Jun, H., et~al.
\newblock Evaluating large language models trained on code.
\newblock \emph{arXiv preprint arXiv:2107.03374}, 2021.

\bibitem[Damiand \& Lienhardt(2014)Damiand and Lienhardt]{damiand2014combinatorial}
Damiand, G. and Lienhardt, P.
\newblock \emph{Combinatorial Maps: Efficient Data Structures for Computer Graphics and Image Processing}.
\newblock CRC Press, 2014.

\bibitem[DiPrete et~al.(2024)DiPrete, Garimella, Garcia~Cardona, and Ray]{diprete2024reinforcement}
DiPrete, B.~C., Garimella, R., Garcia~Cardona, C., and Ray, N.
\newblock Reinforcement learning for block decomposition of planar {CAD} models.
\newblock \emph{Engineering with Computers}, 40\penalty0 (5):\penalty0 2807--2817, 2024.
\newblock \doi{10.1007/s00366-023-01940-6}.

\bibitem[Dong et~al.(2023)Dong, Xiong, Goyal, Zhang, Chow, Pan, Diao, Zhang, et~al.]{dong2023raft}
Dong, H., Xiong, W., Goyal, D., Zhang, Y., Chow, W., Pan, R., Diao, S., Zhang, J., et~al.
\newblock {RAFT}: Reward ranked finetuning for generative foundation model alignment.
\newblock \emph{arXiv preprint arXiv:2304.06767}, 2023.

\bibitem[Eppstein(1999)]{eppstein1999linear}
Eppstein, D.
\newblock Linear complexity hexahedral mesh generation.
\newblock \emph{Computational Geometry}, 12\penalty0 (1--2):\penalty0 3--16, 1999.
\newblock \doi{10.1016/S0925-7721(98)00032-7}.

\bibitem[Fogg et~al.(2018)Fogg, Sun, Makem, Armstrong, and Robinson]{fogg2018singularities}
Fogg, H.~J., Sun, L., Makem, J.~E., Armstrong, C.~G., and Robinson, T.~T.
\newblock Singularities in structured meshes and cross-fields.
\newblock \emph{Computer-Aided Design}, 105:\penalty0 11--25, 2018.
\newblock \doi{10.1016/j.cad.2018.06.002}.

\bibitem[Gao et~al.(2019)Gao, Shen, and Panozzo]{gao2019feature}
Gao, X., Shen, H., and Panozzo, D.
\newblock Feature preserving octree-based hexahedral meshing.
\newblock \emph{Computer Graphics Forum}, 38\penalty0 (5):\penalty0 135--149, 2019.
\newblock \doi{10.1111/cgf.13795}.

\bibitem[Geuzaine \& Remacle(2009)Geuzaine and Remacle]{geuzaine2009gmsh}
Geuzaine, C. and Remacle, J.-F.
\newblock Gmsh: A 3-{D} finite element mesh generator with built-in pre- and post-processing facilities.
\newblock \emph{International Journal for Numerical Methods in Engineering}, 79\penalty0 (11):\penalty0 1309--1331, 2009.

\bibitem[Gregson et~al.(2011)Gregson, Sheffer, and Zhang]{gregson2011allhex}
Gregson, J., Sheffer, A., and Zhang, E.
\newblock All-hex mesh generation via volumetric {PolyCube} deformation.
\newblock \emph{Computer Graphics Forum}, 30\penalty0 (5):\penalty0 1407--1416, 2011.
\newblock \doi{10.1111/j.1467-8659.2011.02015.x}.

\bibitem[Guibas \& Stolfi(1985)Guibas and Stolfi]{guibas1985primitives}
Guibas, L. and Stolfi, J.
\newblock Primitives for the manipulation of general subdivisions and the computation of {Voronoi} diagrams.
\newblock \emph{ACM Transactions on Graphics}, 4\penalty0 (2):\penalty0 74--123, 1985.
\newblock \doi{10.1145/282918.282923}.

\bibitem[Gulcehre et~al.(2023)Gulcehre, Paine, Srinivasan, Konyushkova, Weerts, Sharma, Siddhant, Ahern, et~al.]{gulcehre2023rest}
Gulcehre, C., Paine, T.~L., Srinivasan, S., Konyushkova, K., Weerts, L., Sharma, A., Siddhant, A., Ahern, A., et~al.
\newblock Reinforced self-training ({ReST}) for language modeling.
\newblock \emph{arXiv preprint arXiv:2308.08998}, 2023.

\bibitem[Hanocka et~al.(2019)Hanocka, Hertz, Fish, Giryes, Fleishman, and Cohen-Or]{hanocka2019meshcnn}
Hanocka, R., Hertz, A., Fish, N., Giryes, R., Fleishman, S., and Cohen-Or, D.
\newblock {MeshCNN}: A network with an edge.
\newblock \emph{ACM Transactions on Graphics}, 38\penalty0 (4):\penalty0 1--12, 2019.
\newblock \doi{10.1145/3306346.3322959}.

\bibitem[Havrilla et~al.(2024)Havrilla, Du, Raparthy, Nalmpantis, Dwivedi-Yu, Zhuravinskyi, Hambro, Sukhbaatar, et~al.]{havrilla2024teaching}
Havrilla, A., Du, Y., Raparthy, S.~C., Nalmpantis, C., Dwivedi-Yu, J., Zhuravinskyi, M., Hambro, E., Sukhbaatar, S., et~al.
\newblock Teaching large language models to reason with reinforcement learning.
\newblock \emph{arXiv preprint arXiv:2403.04642}, 2024.

\bibitem[Kazemnejad et~al.(2024)Kazemnejad, Aghajohari, Portelance, Sordoni, Reddy, Courville, and Le~Roux]{kazemnejad2024vineppo}
Kazemnejad, A., Aghajohari, M., Portelance, E., Sordoni, A., Reddy, S., Courville, A., and Le~Roux, N.
\newblock {VinePPO}: Unlocking {RL} potential for {LLM} reasoning through refined credit assignment.
\newblock \emph{arXiv preprint arXiv:2410.01679}, 2024.

\bibitem[Li et~al.(2012)Li, Liu, Xu, Wang, and Guo]{li2012allhex}
Li, Y., Liu, Y., Xu, W., Wang, W., and Guo, B.
\newblock All-hex meshing using singularity-restricted field.
\newblock \emph{ACM Transactions on Graphics}, 31\penalty0 (6):\penalty0 1--11, 2012.
\newblock \doi{10.1145/2366145.2366196}.

\bibitem[Lienhardt(1994)]{lienhardt1994ndimensional}
Lienhardt, P.
\newblock N-dimensional generalized combinatorial maps and cellular quasi-manifolds.
\newblock \emph{International Journal of Computational Geometry \& Applications}, 4\penalty0 (3):\penalty0 275--324, 1994.
\newblock \doi{10.1142/S0218195994000173}.

\bibitem[Liu et~al.(2018)Liu, Zhang, Chien, Solomon, and Bommes]{liu2018singularity}
Liu, H., Zhang, P., Chien, E., Solomon, J., and Bommes, D.
\newblock Singularity-constrained octahedral fields for hexahedral meshing.
\newblock \emph{ACM Transactions on Graphics}, 37\penalty0 (4):\penalty0 1--17, 2018.
\newblock \doi{10.1145/3197517.3201344}.

\bibitem[Livesu et~al.(2013)Livesu, Vining, Sheffer, Gregson, and Scateni]{livesu2013polycut}
Livesu, M., Vining, N., Sheffer, A., Gregson, J., and Scateni, R.
\newblock {PolyCut}: Monotone graph-cuts for {PolyCube} base-complex construction.
\newblock \emph{ACM Transactions on Graphics}, 32\penalty0 (6):\penalty0 1--12, 2013.
\newblock \doi{10.1145/2508363.2508388}.

\bibitem[Ludwig et~al.(2023)Ludwig, Tyson, and Campen]{ludwig2023halfedgecnn}
Ludwig, I., Tyson, D., and Campen, M.
\newblock {HalfedgeCNN} for native and flexible deep learning on triangle meshes.
\newblock \emph{Computer Graphics Forum}, 42\penalty0 (5):\penalty0 e14898, 2023.
\newblock \doi{10.1111/cgf.14898}.

\bibitem[Luo et~al.(2026)Luo, Wu, Wang, and Cai]{luo2026lightweight}
Luo, X., Wu, Z., Wang, Y., and Cai, Y.
\newblock Lightweight online reinforcement learning for block decomposition of {CAD} models.
\newblock In \emph{European Conference on Computer Vision (ECCV)}, 2026.

\bibitem[Mar{\'e}chal(2009)]{marechal2009advances}
Mar{\'e}chal, L.
\newblock Advances in octree-based all-hexahedral mesh generation: Handling sharp features.
\newblock In \emph{Proceedings of the 18th International Meshing Roundtable}, pp.\  65--84, 2009.
\newblock \doi{10.1007/978-3-642-04319-2_5}.

\bibitem[Mitchell(1996)]{mitchell1996characterization}
Mitchell, S.~A.
\newblock A characterization of the quadrilateral meshes of a surface which admit a compatible hexahedral mesh of the enclosed volume.
\newblock In \emph{Symposium on Theoretical Aspects of Computer Science (STACS)}, Lecture Notes in Computer Science, pp.\  465--476, 1996.
\newblock \doi{10.1007/3-540-60922-9_38}.

\bibitem[Narayanan \& Persson(2026)Narayanan and Persson]{narayanan2026playing}
Narayanan, A. and Persson, P.-O.
\newblock Playing to par: Reinforcement learning for provably optimal quadrilateral block decompositions.
\newblock \emph{arXiv preprint arXiv:2609.32146}, 2026.

\bibitem[Narayanan et~al.(2024)Narayanan, Pan, and Persson]{narayanan2024learning}
Narayanan, A., Pan, Y., and Persson, P.-O.
\newblock Learning topological operations on meshes with application to block decomposition of polygons.
\newblock \emph{Computer-Aided Design}, 175:\penalty0 103744, 2024.
\newblock \doi{10.1016/j.cad.2024.103744}.

\bibitem[Ng et~al.(1999)Ng, Harada, and Russell]{ng1999policy}
Ng, A.~Y., Harada, D., and Russell, S.
\newblock Policy invariance under reward transformations: Theory and application to reward shaping.
\newblock In \emph{International Conference on Machine Learning}, 1999.

\bibitem[Nieser et~al.(2011)Nieser, Reitebuch, and Polthier]{nieser2011cubecover}
Nieser, M., Reitebuch, U., and Polthier, K.
\newblock {CubeCover} -- parameterization of {3D} volumes.
\newblock \emph{Computer Graphics Forum}, 30\penalty0 (5):\penalty0 1397--1406, 2011.
\newblock \doi{10.1111/j.1467-8659.2011.02014.x}.

\bibitem[Oh et~al.(2018)Oh, Guo, Singh, and Lee]{oh2018self}
Oh, J., Guo, Y., Singh, S., and Lee, H.
\newblock Self-imitation learning.
\newblock In \emph{International Conference on Machine Learning}, pp.\  3878--3887, 2018.

\bibitem[Owen et~al.(2026)Owen, Brown, Chrisochoides, Garimella, Gu, Ledoux, Lei, and Quadros]{owen2026survey}
Owen, S., Brown, N., Chrisochoides, N., Garimella, R., Gu, X., Ledoux, F., Lei, N., and Quadros, R.
\newblock A survey of {AI} methods for geometry preparation and mesh generation in engineering simulation.
\newblock \emph{Engineering with Computers}, 42\penalty0 (5):\penalty0 171, 2026.
\newblock \doi{10.1007/s00366-026-02379-1}.

\bibitem[Pan et~al.(2023)Pan, Huang, Cheng, and Zeng]{pan2023reinforcement}
Pan, J., Huang, J., Cheng, G., and Zeng, Y.
\newblock Reinforcement learning for automatic quadrilateral mesh generation: A soft actor--critic approach.
\newblock \emph{Neural Networks}, 157:\penalty0 288--304, 2023.
\newblock \doi{10.1016/j.neunet.2022.10.022}.

\bibitem[Pietroni et~al.(2022)Pietroni, Campen, Sheffer, Cherchi, Bommes, Gao, Scateni, Ledoux, Remacle, and Livesu]{pietroni2022hex}
Pietroni, N., Campen, M., Sheffer, A., Cherchi, G., Bommes, D., Gao, X., Scateni, R., Ledoux, F., Remacle, J.-F., and Livesu, M.
\newblock Hex-mesh generation and processing: A survey.
\newblock \emph{ACM Transactions on Graphics}, 42\penalty0 (2):\penalty0 1--44, 2022.
\newblock \doi{10.1145/3554920}.

\bibitem[Schneiders(1996)]{schneiders1996grid}
Schneiders, R.
\newblock A grid-based algorithm for the generation of hexahedral element meshes.
\newblock \emph{Engineering with Computers}, 12\penalty0 (3--4):\penalty0 168--177, 1996.
\newblock \doi{10.1007/BF01198732}.

\bibitem[Schulman et~al.(2016)Schulman, Moritz, Levine, Jordan, and Abbeel]{schulman2016high}
Schulman, J., Moritz, P., Levine, S., Jordan, M.~I., and Abbeel, P.
\newblock High-dimensional continuous control using generalized advantage estimation.
\newblock In \emph{International Conference on Learning Representations}, 2016.

\bibitem[Schulman et~al.(2017)Schulman, Wolski, Dhariwal, Radford, and Klimov]{schulman2017proximal}
Schulman, J., Wolski, F., Dhariwal, P., Radford, A., and Klimov, O.
\newblock Proximal policy optimization algorithms.
\newblock \emph{arXiv preprint arXiv:1707.06347}, 2017.

\bibitem[Shao et~al.(2024)Shao, Wang, Zhu, Xu, Song, et~al.]{shao2024deepseekmath}
Shao, Z., Wang, P., Zhu, Q., Xu, R., Song, J., et~al.
\newblock {DeepSeekMath}: Pushing the limits of mathematical reasoning in open language models.
\newblock \emph{arXiv preprint arXiv:2402.03300}, 2024.

\bibitem[Silver et~al.(2017)Silver, Schrittwieser, Simonyan, Antonoglou, Huang, Guez, Hubert, Baker, et~al.]{silver2017mastering}
Silver, D., Schrittwieser, J., Simonyan, K., Antonoglou, I., Huang, A., Guez, A., Hubert, T., Baker, L., et~al.
\newblock Mastering the game of {Go} without human knowledge.
\newblock \emph{Nature}, 550\penalty0 (7676):\penalty0 354--359, 2017.
\newblock \doi{10.1038/nature24270}.

\bibitem[Verhetsel et~al.(2019{\natexlab{a}})Verhetsel, Pellerin, and Remacle]{verhetsel2019finding}
Verhetsel, K., Pellerin, J., and Remacle, J.-F.
\newblock Finding hexahedrizations for small quadrangulations of the sphere.
\newblock \emph{ACM Transactions on Graphics}, 38\penalty0 (4):\penalty0 1--13, 2019{\natexlab{a}}.
\newblock \doi{10.1145/3306346.3323017}.

\bibitem[Verhetsel et~al.(2019{\natexlab{b}})Verhetsel, Pellerin, and Remacle]{verhetsel2019pyramid}
Verhetsel, K., Pellerin, J., and Remacle, J.-F.
\newblock A 44-element mesh of {Schneiders'} pyramid.
\newblock In \emph{Proceedings of the 27th International Meshing Roundtable}, Lecture Notes in Computational Science and Engineering, pp.\  73--87. Springer, 2019{\natexlab{b}}.
\newblock \doi{10.1007/978-3-030-13992-6_5}.

\bibitem[Xu \& Fang(2026)Xu and Fang]{xu2026popo}
Xu, M. and Fang, H.
\newblock Improving reasoning ability via asynchronous on-policy self-distillation under positive rollouts.
\newblock \emph{arXiv preprint arXiv:2605.06650}, 2026.

\bibitem[Yamakawa \& Shimada(2010)Yamakawa and Shimada]{yamakawa2010schneiders}
Yamakawa, S. and Shimada, K.
\newblock 88-element solution to {Schneiders'} pyramid hex-meshing problem.
\newblock \emph{International Journal for Numerical Methods in Biomedical Engineering}, 26\penalty0 (12):\penalty0 1700--1712, 2010.
\newblock \doi{10.1002/cnm.1256}.

\bibitem[Yu et~al.(2025)Yu, Fang, Tong, and Zhang]{yu2025dlpolycube}
Yu, Y., Fang, Y., Tong, H., and Zhang, Y.~J.
\newblock {DL-Polycube}: Deep learning enhanced generalized polycube method for high-quality hexahedral mesh generation and volumetric spline construction.
\newblock \emph{Engineering with Computers}, 41\penalty0 (3):\penalty0 2069--2092, 2025.
\newblock \doi{10.1007/s00366-025-02161-9}.

\bibitem[Yuan et~al.(2023)Yuan, Yuan, Li, Dong, Lu, Tan, Zhou, and Zhou]{yuan2023scaling}
Yuan, Z., Yuan, H., Li, C., Dong, G., Lu, K., Tan, C., Zhou, C., and Zhou, J.
\newblock Scaling relationship on learning mathematical reasoning with large language models.
\newblock \emph{arXiv preprint arXiv:2308.01825}, 2023.

\bibitem[Zelikman et~al.(2022)Zelikman, Wu, Mu, and Goodman]{zelikman2022star}
Zelikman, E., Wu, Y., Mu, J., and Goodman, N.
\newblock {STaR}: Bootstrapping reasoning with reasoning.
\newblock In \emph{Advances in Neural Information Processing Systems}, pp.\  15476--15488, 2022.

\bibitem[Zhang et~al.(2023)Zhang, Dai, Malkin, Courville, Bengio, and Pan]{zhang2023let}
Zhang, D., Dai, H., Malkin, N., Courville, A., Bengio, Y., and Pan, L.
\newblock Let the flows tell: Solving graph combinatorial optimization problems with {GFlowNets}.
\newblock In \emph{Advances in Neural Information Processing Systems}, 2023.

\bibitem[Zhang et~al.(2025)Zhang, Guan, Wang, Tan, and Jiang]{zhang2025reinforcement}
Zhang, S., Guan, Z., Wang, X., Tan, P., and Jiang, H.
\newblock Reinforcement learning based automatic block decomposition of solid models for hexahedral meshing.
\newblock \emph{Computer-Aided Design}, 182:\penalty0 103850, 2025.
\newblock \doi{10.1016/j.cad.2025.103850}.

\end{thebibliography}
\bibliographystyle{icml2026}

\newpage
\appendix
\onecolumn
\section{Floor and ceiling}
\label{app:floor}
The floor $\underline{I}(\Omega)$ is a lower bound on the irregularity $V+E$
of any valid hex mesh of $\Omega$. It follows from a Gauss--Bonnet-type charge
identity on vertex links together with per-corner local bounds
(\citealp{narayanan2026playing} for the planar analogue, and
\citealp{liu2018singularity,fogg2018singularities} for related singularity
constraints). We note that $\underline{I}$ is a lower bound only: a
decomposition that attains it need not exist for a given solid. The ceiling
$c(\Omega)$ bounds the minimum scaled Jacobian from above. Both are computed
from the input's angles as follows.

A boundary edge of the input has a dihedral angle $\theta$, measured inside
$\Omega$, and is a \emph{ridge} when $\theta$ differs from $180^\circ$; an
input vertex on three or more ridges is a \emph{corner}. A face corner of
angle $\alpha$ asks for $\rho(\alpha)=\max(1,\mathrm{round}(\alpha/90^\circ))$
quadrilaterals and a ridge for $\kappa(\theta)=\max(1,\mathrm{round}(\theta/90^\circ))$
hexahedra. When a rounding is a tie, one choice is made per input ridge or
face angle at reset, and the floor below is minimised over the choices.

\paragraph{Targets.} An interior edge targets $4$ faces and a boundary edge
$\kappa+1$ ($3$ on a flat face). An interior vertex targets $6$ edges, a vertex
on a flat face $5$, a vertex inside a ridge $\kappa+3$, and a corner $c$ with
$r_c$ ridges
\begin{equation}
  t_c = \tfrac12\big(\beta_c + K_c - 2r_c + 6\big),\qquad
  \beta_c=\textstyle\sum_{\text{sectors}}\rho,\quad K_c=\sum_{\text{ridges at }c}\kappa .
  \label{eq:cornertarget}
\end{equation}
These are the valences of a regular grid near each feature; \eqref{eq:cornertarget}
is the corner value consistent with the face angles at the corner (when
$\beta_c+K_c$ is odd no integer target is, and either neighbour is allowed).

\paragraph{Charge identity.} Let $\delta_v$ and $\varepsilon_e$ be the signed
vertex and edge valence deviations of a hex mesh, so $V=\sum_v|\delta_v|$ and
$E=\sum_e|\varepsilon_e|$. At a boundary vertex with $b_v$ boundary edges let
$s_v=b_v-b^*_v$, with $b^*_v=2t_v-6+2r_v-K_v$ the boundary valence of a
regular link ($s_v=0$ inside). Counting the edges and faces of the vertex
link gives, at every vertex,
\begin{equation}
  s_v = 2\delta_v - \textstyle\sum_{e\ni v}\varepsilon_e .
  \label{eq:charge}
\end{equation}
Summing over the boundary quad surface,
$D=\sum_{v}s_v=\sum_c(4-b^*_c)-4\chi(\partial\Omega)$ depends only on the input
(with the targets above, only on its face angles), and summing
\eqref{eq:charge} over all vertices gives the conservation law
$\sum_v\delta_v-\sum_e\varepsilon_e=D/2$. So $V+E\equiv D/2 \pmod 2$.

\paragraph{Floor.} Write $V+E=\sum_v c_v$ with
$c_v=|\delta_v|+\frac12\sum_{e\ni v}|\varepsilon_e|$. By \eqref{eq:charge},
$|s_v|\le 2c_v$, so $V+E\ge|D|/2$. A sharper bound prices each vertex by its
type (flat, ridge with its $\kappa$, or the specific corner): $\gamma_\tau(s)$
is a lower bound on $c_v$ over all vertex links of type $\tau$ with surface
deviation $s$ that a valid mesh allows, computed by enumerating links up to a
size cutoff and bounding larger ones by a counting argument.
The \emph{knapsack floor}
\begin{equation}
  \Gamma(D)=\min\Big\{\textstyle\sum_{v\in\partial\Omega}\gamma_{\tau(v)}(s_v) :
  \sum_v s_v = D\Big\},
\end{equation}
with each corner used once and flat and ridge vertices as often as needed
($\gamma(0)=0$ for them), bounds $V+E$ for every valid mesh. The floor
$\underline I(\Omega)$ is the least integer $\ge\Gamma(D)$ with the parity of
$D/2$; when the enumeration does not cover a corner we fall back to $|D|/2$.
The floor is a bound, not always attained.

\paragraph{Ceiling.} For unit vectors $u_1,u_2,u_3$ at a hex corner,
$\det(u_1,u_2,u_3)=\sin a_{12}\sin a_{13}\sin\theta_1$, with $a_{ij}$ the
angles between them and $\theta_1$ the dihedral along $u_1$. So a corner's
scaled Jacobian is at most the sine of each of its face angles and dihedrals.
For a mesh at target on every input ridge and corner this gives three
bounds: a ridge of dihedral $\theta$ split among $\kappa$ hexahedra gives
$\sin(\theta/\kappa)$; a corner's $b^*_c$ boundary quadrilaterals split over
its sectors gives the best $\min_i\sin(\alpha_i/n_i)$ over splits $\sum_i
n_i=b^*_c$ (each sector of angle $\alpha_i$ holding at least
$\lfloor\alpha_i/180^\circ\rfloor+1$); and a corner with three ridges and
target $3$ is a single hex corner, whose scaled Jacobian is the determinant of
the three unit ridge directions. The ceiling $c(\Omega)$ is the least of these
over the input.

\section{Restart search}
\label{app:search}
Mining runs restart search with the previous round's policy at $\beta=3$ and one
sampling seed per round, with a budget of $6{,}000$ committed moves per solid, the same in every round and for both seeds.
Each attempt is an ordinary episode under the termination rules of
\cref{sec:mdp:term}, at the training threshold $\bar g=0.5$: it ends with
success, by defect blow-up, at the pick budget, when no legal move remains, or on a sampling error.
A failed attempt is discarded, its committed moves count against the budget,
and the next attempt starts from the raw solid. There is no other abort rule
and no rewinding. Over the $62{,}021$ searches of both seeds, $702{,}286$ of
the $702{,}981$ failed attempts ($99.9\%$) ended by defect blow-up.

\section{Rewards}
\label{app:reward}
Expert iteration and POPO use no reward. The policy-gradient arms of
Section~\ref{sec:exp:pg} train on a dense, potential-based reward that the
environment computes per move, at the threshold $\bar g=0.4$. With $F$, $C$, $V$, $E$ as in Section~\ref{sec:mdp:term},
floor $\underline{I}$, and $q$ the minimum scaled Jacobian over the cells that
are already hexahedra ($q=1$ if there are none), the \emph{defect} of a state is
\begin{equation}
  \mathcal{D}(s) = F + C + \tfrac14\max\!\big(0,\ V+E-\underline{I}\big)
                   + 4\max\!\big(0,\ g-q\big),
  \label{eq:defect}
\end{equation}
with $g$ the quality threshold \eqref{eq:gate}. The reward is potential-based
\citep{ng1999policy}, with potential $\Phi(s)=-\mathcal{D}(s)/S$ normalised by
$S=\max(\mathcal{D}(s_0),\,2E_{\mathrm{in}})$:
\begin{equation}
  r_t = \Phi(s_{t+1})-\Phi(s_t) + \mathbf{1}[\text{accepted}] - \frac{0.1}{T},
  \label{eq:reward}
\end{equation}
undiscounted, with $T$ as in Section~\ref{sec:mdp:term}. A decision's reward is
the sum of its picks' rewards. The weights $\tfrac14$ and $4$ were set by hand
and not tuned. GRPO normalises the episode return within each solid's $16$
episodes. PPO fits a separate critic, an MLP with two hidden layers of $256$
units on the policy's context vector (Section~\ref{sec:policy:trunk}), to the
return-to-go, and takes generalised advantage estimates from it. VinePPO
estimates the value at $11$ points of an episode as the mean return-to-go of
$4$ Monte Carlo continuations. None of the arms uses the network's own value
head.

\section{Recipe rounds}
\label{app:recipe}
\begin{table}[h]
\caption{Recipe rounds on validation, per seed, at each round's reporting
$\beta$, quality threshold $\bar g=0.4$. $V$: the stopping score, pooled $\mathbb{E}[\mathrm{b20}]$ over the
small- and large-chamfered selection sets (Small sel., Large sel.); bold marks
$\pi_R$. Entries are $\mathbb{E}[\mathrm{b20}]$ / per-sample. Small sel.: $20$
samples per solid; Large sel.: $10$ samples, plug-in estimate; the
small-chamfered validation set (Small val.; $60$ samples; $\pi_0$, $\pi_R$ and
seed 1's round 3 only) is never used for any choice. New: solids added to the archive in the round (in round 1, the whole
archive); Upg.: archived solids whose solution was replaced by a better one.}
\label{tab:recipe}
\centering
\footnotesize
\setlength{\tabcolsep}{3pt}
\resizebox{\columnwidth}{!}{%
\begin{tabular}{@{}cccrrcccc@{}}
\toprule
Seed & Round & $\beta$ & New & Upg. & $V$ & Small sel. & Large sel. & Small val. \\
\midrule
1 & $0$ & 4 & ---     & ---  & 0.483 & 0.974 / 0.595 & 0.152 / 0.027 & 0.947 / 0.534 \\
  & $1$ & 4 & 7{,}094 & ---  & 0.533 & 0.974 / 0.664 & 0.236 / 0.063 & --- \\
  & $2$ & 4 & 896     & 291  & 0.537 & 0.974 / 0.710 & 0.242 / 0.071 & --- \\
  & $3$ & 3 & 421     & 445  & 0.543 & 0.965 / 0.657 & 0.259 / 0.060 & 0.972 / 0.584 \\
  & $4$ & 4 & 213     & 564  & 0.555 & 0.974 / 0.718 & 0.272 / 0.077 & --- \\
  & $5$ & 3 & 164     & 713  & \textbf{0.574} & 0.991 / 0.707 & 0.293 / 0.080 & 0.987 / 0.634 \\
\midrule
2 & $0$ & 4 & ---     & ---  & 0.503 & 0.956 / 0.628 & 0.198 / 0.044 & 0.961 / 0.548 \\
  & $1$ & 4 & 7{,}259 & ---  & 0.528 & 0.991 / 0.674 & 0.216 / 0.058 & --- \\
  & $2$ & 3 & 770     & 283  & 0.537 & 0.965 / 0.636 & 0.248 / 0.054 & --- \\
  & $3$ & 3 & 346     & 440  & \textbf{0.567} & 0.991 / 0.699 & 0.281 / 0.070 & 0.980 / 0.617 \\
  & $4$ & 3 & 263     & 581  & 0.567 & 0.982 / 0.727 & 0.286 / 0.084 & --- \\
  & $5$ & 3 & 227     & 736  & 0.544 & 0.974 / 0.698 & 0.254 / 0.070 & --- \\
\bottomrule
\end{tabular}}
\end{table}

\section{Update rules at matched compute}
\label{app:pg}
\begin{table}[h]
\caption{Update rules at matched rollouts and optimizer steps, from seed 1's
$\pi_0$; three iterations on $64$ small chamfered solids, threshold $\bar g=0.4$,
$\beta=3$. Wins: won episodes of the $1{,}024$ sampled for iterations 2 and 3
($\pi_0$: $339$ for iteration 1). Train and held-aside: per-sample solve rate
on the trained and the held-aside $64$ solids ($16$ samples each). Small
val.: the small-chamfered validation set, $40$ samples per solid; Zero: solids
never solved, of $200$. ``$+$KL'': with the KL guard. The policy-gradient arms
train on the shaped reward of Appendix~\ref{app:reward}. Bold: significantly above
$\pi_0$ (the $95\%$ interval of the paired difference excludes zero;
\cref{fig:pg}); no arm is significantly above $\pi_0$ on the held-aside solids
or on validation.}
\label{tab:pg}
\centering
\footnotesize
\setlength{\tabcolsep}{3pt}
\begin{tabular}{@{}lccccccc@{}}
\toprule
 & \multicolumn{2}{c}{Wins} & \multicolumn{2}{c}{Pool ps} & \multicolumn{3}{c}{Small val.} \\
\cmidrule(lr){2-3}\cmidrule(lr){4-5}\cmidrule(l){6-8}
Arm & it.~2 & it.~3 & Train & Held & ps & $\mathbb{E}[\mathrm{b20}]$ & Zero \\
\midrule
$\pi_0$            & --- & --- & 0.313 & 0.360 & 0.513 & 0.944 & 6 \\
POPO, $\pi/Z_+$    & 362 & 375 & \textbf{0.388} & 0.412 & 0.500 & 0.906 & 13 \\
POPO, uniform      & 350 & 330 & 0.311 & 0.332 & 0.474 & 0.943 & 6 \\
PPO$+$KL           & 242 & 144 & 0.166 & 0.183 & 0.354 & 0.870 & 17 \\
VinePPO$+$KL       & 291 & 164 & 0.100 & 0.140 & 0.271 & 0.798 & 26 \\
GRPO$+$KL          & 287 & 224 & 0.181 & 0.227 & 0.398 & 0.863 & 20 \\
PPO                & 98  & 24  & 0.006 & 0.024 & --- & --- & --- \\
GRPO               & 71  & 5   & 0.004 & 0.011 & --- & --- & --- \\
\bottomrule
\end{tabular}
\end{table}

\section{Certified data}
\label{app:lineage}

\begin{table}[h]
\caption{Certified clone data $\mathcal{P}_0$ of the final recipe. Every source is
re-collected with held-out exclusion (\emph{Excl.}: solids it dropped). CC: Catmull--Clark. Small
certified solids are replayed twice, in independent random orders, and the
others once. A replay fails when the collector cannot reproduce the plan's
accepted end state; failed replays are discarded. Within $\mathcal{P}_0$, the
large grid certificates have weight $0.5$ and every other source has weight $1$.
Cells: grid cells of the certificate.}
\label{tab:p0}
\centering
\small
\setlength{\tabcolsep}{4pt}
\begin{tabular}{@{}llrrrr@{}}
\toprule
Source & Contents & Solids & Excl. & Replays ok / failed & States \\
\midrule
\multicolumn{6}{@{}l}{\emph{Grid certificates}} \\
Small certified    & polycube, tilted, Geo3D, polar, CC; $\le 16$ cells & 5{,}000 & 357 & 9{,}236 / 50 & 477{,}549 \\
Small certified, short & same families, fresh draws & 5{,}000 & 352 & 9{,}261 / 35 & 475{,}989 \\
Large Geo3D        & $30$--$60$ cells & 1{,}000 & 0 & 957 / 43 & 191{,}061 \\
Large polycube / tilted & $30$--$60$ cells & 3{,}000 & 0 & 2{,}941 / 59 & 517{,}040 \\
Geo3D, long loops  & $17$--$60$ cells, face-loop cap $24$ & 2{,}000 & 1 & 1{,}920 / 79 & 335{,}326 \\
\midrule
\multicolumn{6}{@{}l}{\emph{Chamfer plans}} \\
Edge chamfer (v1)  & grid + edge-chamfer templates & 644 & 0 & 638 / 6 & 43{,}709 \\
Chamfer (v2)       & edge and corner chamfers, natural mix & 701 & 0 & 686 / 15 & 76{,}176 \\
Corner chamfer (v2) & corner chamfers only & 1{,}810 & 0 & 1{,}777 / 33 & 244{,}005 \\
\midrule
\multicolumn{6}{@{}l}{\emph{Swept prisms}} \\
Pilot              & 2D quad meshes, extruded & 222 & 0 & 218 / 4 & 15{,}363 \\
Straight           & & 2{,}242 & 0 & 2{,}194 / 48 & 152{,}784 \\
Big                & $16$--$24$ corners & 437 & 0 & 416 / 21 & 40{,}104 \\
Big, chamfered     & $16$--$24$ corners, concave-heavy & 409 & 0 & 380 / 29 & 42{,}126 \\
\midrule
Total              & & 22{,}465 & 710 & & 2{,}611{,}232 \\
\bottomrule
\end{tabular}
\end{table}
The chamfer and prism collections ran against an earlier exclusion list; a
re-check of their instance files against the full list of $1{,}757$ held-out
signatures finds no overlap.

Each recipe round trains for one epoch on two groups
(\cref{tab:mix}). The archive is re-collected, every kept solution replayed
once in a random order, at weight $1$. $\mathcal{P}_0$ is subsampled to $25\%$
of the round's states, keeping its internal weights; its sampling factor
therefore grows with the archive, from $0.14$ in round 1 to $0.22$ in round 5
($0.07$ to $0.11$ for the large grid certificates). Mining draws from one
fixed set of $12{,}000$ solids, $3{,}000$ in each of four classes, the same
for both seeds and every round.

\begin{table}[h]
\caption{Training mixture per seed and round. Archived solids by mining class:
SC small chamfered, CC concave chamfered, LC large chamfered, RP
rectilinear; Sol.: kept solutions (at most two per solid). States: training
states from the archive and from $\mathcal{P}_0$ (after subsampling), and the
states trained on in the round's single epoch (whole batches of $64$).}
\label{tab:mix}
\centering
\footnotesize
\setlength{\tabcolsep}{3pt}
\begin{tabular}{@{}ccrrrrrrrr@{}}
\toprule
& & \multicolumn{5}{c}{Archived solids} & \multicolumn{3}{c}{States} \\
\cmidrule(lr){3-7}\cmidrule(l){8-10}
Seed & Round & SC & CC & LC & RP & Sol. & Archive & $\mathcal{P}_0$ & Trained \\
\midrule
1 & 1 & 2{,}872 & 983     & 390     & 2{,}849 & 7{,}094 & 963{,}684   & 321{,}228 & 1{,}284{,}864 \\
  & 2 & 2{,}938 & 1{,}436 & 662     & 2{,}954 & 8{,}265 & 1{,}192{,}148 & 397{,}383 & 1{,}589{,}504 \\
  & 3 & 2{,}958 & 1{,}640 & 834     & 2{,}979 & 8{,}979 & 1{,}331{,}856 & 443{,}952 & 1{,}775{,}744 \\
  & 4 & 2{,}968 & 1{,}744 & 927     & 2{,}985 & 9{,}455 & 1{,}424{,}698 & 474{,}899 & 1{,}899{,}584 \\
  & 5 & 2{,}974 & 1{,}823 & 1{,}002 & 2{,}989 & 9{,}904 & 1{,}500{,}295 & 500{,}098 & 2{,}000{,}384 \\
\midrule
2 & 1 & 2{,}885 & 1{,}038 & 483     & 2{,}853 & 7{,}259 & 972{,}507   & 324{,}169 & 1{,}296{,}640 \\
  & 2 & 2{,}938 & 1{,}411 & 728     & 2{,}952 & 8{,}300 & 1{,}169{,}989 & 389{,}996 & 1{,}559{,}936 \\
  & 3 & 2{,}955 & 1{,}588 & 865     & 2{,}967 & 8{,}925 & 1{,}284{,}034 & 428{,}011 & 1{,}712{,}000 \\
  & 4 & 2{,}964 & 1{,}720 & 978     & 2{,}976 & 9{,}462 & 1{,}383{,}335 & 461{,}112 & 1{,}844{,}416 \\
  & 5 & 2{,}969 & 1{,}831 & 1{,}083 & 2{,}982 & 9{,}965 & 1{,}484{,}037 & 494{,}679 & 1{,}978{,}688 \\
\bottomrule
\end{tabular}
\end{table}

\section{Compute}
\label{app:compute}
\begin{table}[h]
\caption{Compute per seed and round, from the cluster's accounting: GPU-hours
(A5000) and allocated CPU core-hours (allocated cores $\times$ wall time).
Val.: the validation evaluations used to pick $\beta$ and to stop. Held-out
evaluation is excluded ($581$--$639$ core-hours per battery). Collecting
$\mathcal{P}_0$ takes a further $401$ core-hours, shared by both seeds.
Failed attempts that used time are included. $^\ast$On one A40.}
\label{tab:compute}
\centering
\footnotesize
\setlength{\tabcolsep}{4pt}
\begin{tabular}{@{}ccrrrrr@{}}
\toprule
Seed & Round & GPU-h & Mining & Archive & Val. & CPU total \\
\midrule
1 & $\pi_0$ & 15.7 & ---    & ---  & 376 & 439 \\
  & 1       & 1.8  & 2{,}352 & 243 & 263 & 2{,}872 \\
  & 2       & 2.2  & 1{,}880 & 295 & 189 & 2{,}375 \\
  & 3       & 2.3  & 1{,}718 & 324 & 191 & 2{,}245 \\
  & 4       & 2.7$^\ast$ & 1{,}381 & 332 & 143 & 1{,}880 \\
  & 5       & 2.7$^\ast$ & 1{,}415 & 360 & 204 & 2{,}003 \\
\midrule
2 & $\pi_0$ & 15.7 & ---    & ---  & 407 & 470 \\
  & 1       & 1.8  & 2{,}357 & 247 & 183 & 2{,}796 \\
  & 2       & 2.0  & 1{,}894 & 282 & 194 & 2{,}381 \\
  & 3       & 2.3  & 1{,}760 & 307 & 184 & 2{,}262 \\
  & 4       & 2.4  & 1{,}389 & 336 & 247 & 1{,}984 \\
  & 5       & 2.6  & 1{,}385 & 370 & 193 & 1{,}961 \\
\bottomrule
\end{tabular}
\end{table}
CPU total also counts the clone's and each round's training-job CPUs and the
small data-mixing steps. The seed-1 round-1 and seed-2 round-4 validation
counts include the $60$-sample run on the small-chamfered validation set made
with the held-out battery.

\section{Boundary-topology filter}
\label{app:neardup}
We compare every test solid with every solid in the source files of
$\mathcal{P}_0$, taken before the collection filter, so the check is
conservative. Two solids match when their boundary incidence graphs (vertices,
edges and faces, each face labelled by its number of sides) have the same
Weisfeiler--Lehman hash; geometry, feature sizes and orientation are ignored.
For the unseen-corner solids we confirm by exact isomorphism that none matches.
\Cref{tab:neardup} gives the counts. Before the filter, solids of the small
chamfered set that are axis-aligned stretches of one another were reduced to
one each, which leaves $278$ of $286$; the other generated sets had no such
pairs. Of the seven unseen-corner solids we keep one square pyramid and drop
three stretched or rotated copies of it, which leaves four.
\begin{table}[h]
\caption{Test solids drawn, matched in boundary topology by a solid of the
cloning data $\mathcal{P}_0$, and reported. Matches with the expert-iteration
archives are not removed.}
\label{tab:neardup}
\centering
\footnotesize
\setlength{\tabcolsep}{4pt}
\begin{tabular}{@{}lrrr@{}}
\toprule
Set & Drawn & Matched in $\mathcal{P}_0$ & Reported \\
\midrule
Rectilinear         & 472 & 212 & 260 \\
Small chamfered     & 278 & 178 & 100 \\
Large chamfered     & 199 & 6   & 193 \\
Unseen corner types & 4   & 0   & 4   \\
\bottomrule
\end{tabular}
\end{table}

\section{Implementation details}
\label{app:impl}

\paragraph{Masks.}
A face insert's loop is grown by a bounded depth-first search over the
volume's edge graph, and only continuations from which the loop can still
close are offered. A closing step that would leave either half with fewer than
four faces is rejected, and loops are limited to $24$ edges. A seed found to
have no closable loop is discarded, the decision is redrawn, and the seed is
removed from the candidate set until the mesh changes. The pick budget of
\cref{sec:mdp:term} counts elementary picks, not decisions.

\paragraph{Untangling.}
The untangling pass at the acceptance test runs Gauss--Seidel sweeps with
Newton steps on the cell Jacobians.

\paragraph{Observation.}
The $23$ per-dart features are: the target and actual valences at the dart's
vertex, its head vertex and its edge; the number of sides of its face, faces of
its cell, edges of its cell and cells at its vertex; the length of its edge
divided by the median edge length in the window; the corner angles of its face
at both endpoints and the dihedral angle of its cell at its edge, in multiples
of $90^\circ$; boundary flags for its edge and face; a one-hot class of its
vertex relative to the input (corner, ridge, surface, interior); and three
flags for an insert in progress. Angles use Newell normals, so they are
defined on non-planar faces. The ten global scalars are $V$, $E$, $F$, $C$, the
number of odd-sided faces and the floor, each relative to the initial defect;
the number of cells relative to the input's face count; the mesh size relative
to the window; the worst scaled Jacobian of the finished hexahedra; and the
rejection flag.

\paragraph{Edge-insert pair features.}
In \cref{eq:pair}, $\phi_{uv}$ holds seven combinatorial features of the split
(the face's size, the sizes of the two parts and whether each is a
quadrilateral) and $\gamma_{uv}$ eight geometric features of $v$ seen from $u$
in a frame attached to $u$'s face (relative position and distance in units of
the median edge length, angles to the face's edges at both ends, and whether
the new edge continues an existing one). The residual MLP is zero-initialised,
so training starts from the endpoint prior.

\paragraph{Parameters and optimiser.}
Of the policy's $0.46$M parameters, $0.37$M are in the trunk, $0.05$M in the
context network and per-dart head, and $0.04$M in the pair-scoring MLP. We use
Adam with learning rate $5\times10^{-5}$ for the trunk and $3\times10^{-4}$ for
the heads, batches of $64$ states, gradient clipping at $1$, and a
$150{,}000$-sample shuffle buffer over the streamed dataset.

\paragraph{Evaluation.}
Realised b20 from a single block of $20$ samples is noisy: on the small
chamfered set its standard deviation over random 20-of-60 subsets is
$0.013$--$0.019$, and the three blocks of seeds $1$--$20$, $21$--$40$ and
$41$--$60$ lie up to $1.0$ standard deviation from the mean ($1.6$ on the
rectilinear and large chamfered sets). The reporting $\beta$ maximises
$\mathbb{E}[\mathrm{b20}]$ pooled over the two selection sets weighted by
solid count, and ties within $0.005$ go to $\beta=3$.

\end{document}